\documentclass[letterpaper,11pt]{article}
\usepackage{tabularx, ragged2e, caption, lipsum}
\usepackage{amsmath}  
\usepackage{graphicx} 
\usepackage[margin=1in,letterpaper]{geometry} 
\usepackage{cite} 
\usepackage[final]{hyperref} 
\usepackage{adjustbox}
\usepackage{multirow}
\usepackage{siunitx}
\usepackage{array}
\newcolumntype{P}[1]{>{\centering\arraybackslash}p{#1}}

\begin{document}

\title{Diversity in Faces}
\author{Michele Merler, Nalini Ratha, Rogerio Feris, John R. Smith\\IBM Research AI @ IBM T. J. Watson Research Center\\Yorktown Heights, NY 10598, USA\\Contact: jsmith@us.ibm.com}
\maketitle

\begin{abstract}
Face recognition is a long-standing challenge in the field of Artificial Intelligence (AI).  The goal is to create systems that detect, recognize, verify and understand characteristics of human faces.  There are significant technical hurdles in making these systems accurate, particularly in unconstrained settings, due to confounding factors related to pose, resolution, illumination, occlusion and viewpoint.  However, with recent advances in neural networks, face recognition has achieved unprecedented accuracy, built largely on data-driven deep learning methods.  While this is encouraging, a critical aspect limiting face recognition performance in practice is intrinsic facial diversity.  Every face is different.  Every face reflects something unique about us.  Aspects of our heritage -- including race, ethnicity, culture, geography -- and our individual identity -- age, gender and visible forms of self-expression -- are reflected in our faces.  Faces are personal.  We expect face recognition to work accurately for each of us.  Performance should not vary for different individuals or different populations.  As we rely on data-driven methods to create face recognition technology, we need to answer a fundamental question: does the training data for these systems fairly represent the distribution of faces we see in the world?  At the heart of this core question are deeper scientific questions about how to measure facial diversity, what features capture intrinsic facial variation and how to evaluate coverage and balance for face image data sets.  Towards the goal of answering these questions, Diversity in Faces ($DiF$) provides a new data set of annotations of one million publicly available face images for advancing the study of facial diversity.  The annotations are generated using ten facial coding schemes that provide human-interpretable quantitative measures of intrinsic facial features.  We believe that making these descriptors available will encourage deeper research on this important topic and accelerate efforts towards creating more fair and accurate face recognition systems. 

\end{abstract}

\section{Introduction} \label{sec:intro}

Have you ever been treated unfairly? How did it make you feel? Probably not too good. Most people generally agree that a fairer world is a better world.  Artificial Intelligence (AI) has enormous potential to make the world a better place.  Yet, as we develop and apply AI towards addressing a broad set of important challenges, we need to make sure the AI systems themselves are fair and accurate.  Recent advances in AI technology have produced remarkable capabilities for accomplishing sophisticated tasks, like translating speech across languages to augment communications and bridge cultures, improving complex interactions between people and machines, and automatically recognizing contents of video to assist in safety applications.  Much of the recent power of AI comes from the use of data-driven deep learning to train increasingly accurate models by using growing amounts of data.  However, the strength of these techniques can also be their inherent weakness.  These AI systems learn what they are taught.  If they are not taught with robust and diverse data sets, accuracy and fairness are at risk.  For that reason, AI developers and the research community need to be thoughtful about what data they use for training. This is essential for developing AI systems which can help to make the world more fair.

The challenge in training AI systems is manifested in a very apparent and profound way with face recognition technology. Today, there can be difficulties in making face recognition systems that meet fairness expectations. The heart of the problem lies not with the AI technology itself, per se, but with how the systems are trained.  For face recognition to perform as desired –- to be both accurate and fair –-  training data must provide sufficient balance and coverage.  The training data sets should be large enough and diverse enough to learn the many ways in which faces inherently differ. The images must reflect the diversity of features in faces we see in the world.  This raises the important question of how we measure and ensure diversity for faces.  On one hand, we are familiar with how faces may differ according to age, gender and skin color. But, as prior studies have shown, these dimensions are inadequate for characterizing the full range of diversity of faces. Dimensions like face symmetry, facial contrast, and the sizes, distances and ratios of the various attributes  of the face (eyes, nose, forehead, etc.), among many others, are important. 
 
Diversity in Faces ($DiF$) is a new large and diverse data set designed to advance the study of fairness and accuracy in face recognition technology.  $DiF$ provides a data set of annotations of one million face images.  The $DiF$ annotations are made on faces sampled from the publicly available YFCC-100M data set of 100 million images~\cite{yahoo100m_CACM16}.  The $DiF$ data set provides a comprehensive set of annotations of intrinsic facial features that includes craniofacial distances, areas and ratios, facial symmetry and contrast, skin color, age and gender predictions, subjective annotations, and pose and resolution.  The facial coding schemes, summarized in Table~\ref{tab:coding_schemes}, are among the strongest identified in the scientific literature and build a solid foundation to our collective knowledge.  We believe that making these descriptors available will help accelerate the study of diversity and coverage of data towards creating more fair and accurate face recognition systems. 

\begin{table}[ht]
\begin{tabular}[ht!]{|m{.2in}|m{1.7in}|m{4in}|} \hline
  \bf \# & \bf Facial Coding Scheme & \bf Reference \\ \hline 
    \hline
  1 &  Craniofacial Distances & L. G. Farkas, {\it Anthropometry of the Head and Face}, Raven Press, 1994 \cite{anthropometry_book94}. \\
    \hline
  2 &   Craniofacial Areas & L. G. Farkas, et. al, ``International anthropometric study of facial morphology in various ethnic groups/races,'' {\it J Craniofac Surg.} 2005 Jul;16(4), pp. 615-46 \cite{anthropmetry05}.\\
    \hline
  3 &    Craniofacial Ratios&N. Ramanathan, R. Chellappa, ``Modeling Age Progression in Young Faces,'' {\it Intl. Conf. on Computer Vision and Pattern Recognition (CVPR)}, 2006, pp. 387-394 \cite{face_age_modeling06}. \\
    \hline
  4 &  Facial Symmetry & A. C. Little, B. C. Jones, L. M. DeBruine, ``Facial attractiveness: evolutionary based research,'' {\it Philos Trans R Soc Lond B Biol Sci.} 2011 Jun 12;366(1571), pp. 1638-59 \cite{facial_attractiveness11}. \\
    \hline
  5 &  Facial Contrast & A. Porcheron, E. Mauger, R. Russell, ``Aspects of Facial Contrast Decrease with Age and Are Cues for Age Perception,'' {\it PLoS One} 8(3), Mar. 6, 2013 \cite{facial_contrast17}. \\
    \hline
  6 &  Skin Color & A. Chardon I. Cretois and C. Hourseau, ``Skin colour typology and suntanning pathways,'' {\it Intl. Journal of Cosmetic Science}, Aug. 1991, 13(4), pp. 191-208 \cite{ita_ijcs1991}. \\
   \hline 
 7 &  Age  & R. Rothe, R. Timofte, L. Van Gool, ``Deep Expectation of Real and Apparent Age from a Single Image Without Facial Landmarks," {\it L. Int J Comput Vis} (2018) 126: 144 \cite{age_image18}. \\
   \hline
 8 &   Gender & Same as above\\
   \hline
 9 & Subjective Annotation & Z. Liu, P. Luo, X. Wang, X. Tang, ``Deep Learning Face Attributes in the Wild,'' {\it IEEE Intl. Conf. on Computer Vision (ICCV)}, 2015 \cite{CelebA_ICCV15}  \\ 
 \hline 
  10 & Pose and Resolution & X. Zhu, D. Ramanan, ``Face Detection, Pose Estimation, and Landmark Localization in the Wild,'' {\it Intl. Conf. on Computer Vision and Pattern Recognition (CVPR)}, 2012 \cite{face_detection_cvpr12}. \\
   \hline
   \end{tabular}
\caption{Summary of the ten facial coding schemes used in the $DiF$ data set and their references.}\label{tab:coding_schemes}
\end{table}

In this paper, we describe the development and analysis of the $DiF$ data set.  The paper is organized as follows: in Section~\ref{sec:related}, we review the state of face recognition technology and examine how different face image data sets are used today.  We discuss some of the shortcomings from over-reliance on narrow data sets. In Section~\ref{sec:data set}, we describe the process for creating the $DiF$ data set. In Section~\ref{sec:codingschemes}, we describe the implementation of the ten facial coding schemes.  In Section~\ref{sec:analysis}, we provide a statistical analysis of the coding schemes extracted for the face images.  In Section~\ref{sec:conclusion}, we summarize and discuss future directions.

\section{Related Work} \label{sec:related}

Face recognition is a long-standing topic in computer vision, and AI broadly.  Computer-based face recognition was addressed as far back as the 1970s with Takeo Kanade`s seminal thesis on recognizing faces using a set of manually defined points corresponding to nose, mouth, eyes and other features.  Modest by today's standards, his work processed $800$ photographs and conducted experiments involving identification of 20 people~\cite{kanade73}.  Two decades later, a  significant development came from Matthew Turk and Alex Pentland, who developed an appearance-based technique called {\em eigenfaces} that models faces holistically from image data~\cite{turk91}.  This kind of data-driven methodology was subsequently helped by numerous efforts of curating large and growing face image data sets.
The community has built open evaluations around these data sets, such as MegaFace~\cite{Megaface_CVPR16}, MS-Celeb~\cite{MSCELEB1M_ECCV16} and the {NIST} Face Recognition Vendor Test (FRVT)\footnote{https://www.nist.gov/programs-projects/face-recognition-vendor-test-frvt}.  


One prominent example of an early face data set and open evaluation is Labeled Faces in the Wild (LFW), which is comprised of $13,233$ face photos from the Web of $5,749$ individuals, mostly celebrities and public figures, captured in unconstrained conditions of lighting, pose and expression~\cite{LFW_Tech07}.  LFW gained significant focus from the research community upon its release.  Eventually, with the advent of deep learning techniques \cite{krizhevsky2012imagenet,taigman2014deepface}, face recognition performance on LFW reached near-perfect results with $99.8\%$ accuracy \cite{LFWlink,liu2015targeting}.  Megaface defined a follow-on larger data set comprised of faces from $690,572$ unique individuals which was made more difficult with the addition of $1$ million face image distractors~\cite{Megaface_CVPR16}.  Although early results produced low accuracy in the range of $50\%-60\%$, ultimately, performance reached near-perfect levels of $99.9\%$ \cite{megafacelink,deng2018arcface}.  Other data sets and evaluations such as CelebA have brought focus to a wider set of problems in face recognition such as face attribute recognition.  CelebA provides a data set of $202,599$ face images with annotations of $40$ attributes such as `eyeglasses,' `smiling,' and `mustache'~\cite{CelebA_ICCV15}.  State-of-art systems have achieved greater than $90\%$ mean accuracy across the CelebA attribute set and as high as $99\%$ for some attributes. Many other face data sets and evaluations have produced similar improvements using deep learning methods~\cite{MSCELEB1M_ECCV16,VGG2_FG18,VGG_BMVC15,imdbface_ECCV18,age_image18,CASIA_ARXIV14,UMDFaces_ARXIV16,CelebA_ICCV15,CACD_TMM15,IJB-C_ICB18,FaceScrub_ICIP14,IJB-B_CVPRW17,Adience_TIFS14,UTKface_CVPR17,AgeDB_CVPRW17,lfwplus_TPAMI17}.  The healthy progress on face recognition, as measured on these data sets and evaluations, has raised expectations in the technology.

\begin{small}
\begin{table}[ht]
\begin{center}
\begin{adjustbox}{angle=0}
\begin{tabular}{|l|r|r|r|r|r|r|r|}\hline
\multirow{2}{0pt} &   \multicolumn{7}{c|}{\bf Age Group}  \\\hline \hline
                   \bf Data set& \bf 0-3 & \bf 4-12 & \bf 13-19 &  \bf 20-30 & \bf 31-45 & \bf 46-60 & \bf $>$60  \\ \hline \hline 
LFW \cite{LFW_Tech07}	        & 1.0\% &10.6\% & \multicolumn{2}{|c|}{25.4\%} & \multicolumn{2}{c|}{29.6\%} & 33.4\%  \\ \hline
IJB-C \cite{IJB-C_ICB18}        & 0.0\% & 0.0\% & 0.5\% & 16.2\% & 35.5\% & 35.1\% & 12.7\% \\ \hline
Pubfig \cite{Pubfig_PAMI11}     & 1.0\% & 10.8\% & \multicolumn{2}{|c|}{55.5\%} &  \multicolumn{2}{c|}{21.0\%} & 11.7\%  \\ \hline
CelebA \cite{CelebA_ICCV15}     & \multicolumn{5}{c|}{77.8\%} & \multicolumn{2}{c|}{22.1\%}\\ \hline
UTKface \cite{UTKface_CVPR17}   & 8.8\% & 6.5\% & 5.0\% & 33.6\% & 22.6\% & 13.4\% & 10.1\%  \\ \hline
AgeDB \cite{AgeDB_CVPRW17}      & 0.1\% & 0.52\% & 2.7\% & 17.5\% & 31.8\% & 24.5\% & 22.9\%  \\ \hline
IMDB-Face \cite{imdbface_ECCV18}& 0.9\% & 3.5\% & 33.2\% & 36.5\% & 18.8\% & 5.4\% & 1.7\% \\ \hline
\end{tabular}
\end{adjustbox}
\end{center}
\vspace{-0.5cm}
\caption{Distribution of age groups for seven prominent face image data sets.}
\label{tab:sets_distribution_1}
\end{table}
\end{small}

However, high accuracy on these data sets does not readily translate into equivalent accuracy in deployments~\cite{gendershades_FAT18,raji_aeis19}.  The reason is that different or broader distributions of faces, as well as varied environmental conditions, are found in real applications.  Face recognition systems that are trained within only a narrow context of a specific data set will inevitably acquire bias that skews learning towards the specific characteristics of the data set.  This narrow context appears as under-representation or over-representation of certain types of faces in many of the publicly available data sets.  Table~\ref{tab:sets_distribution_1} shows some of the big differences in distribution of age groups for seven prominent face image data sets.  Generally, there is a skew away from younger and older ages.  Some of the differences are quite dramatic.  For example, $36.5\%$ of faces in IMDB-Face are for individuals 20-30 years of age, whereas IJB-C has $16.2\%$ of faces in this age group.  

\begin{small}
\begin{table}[ht]
\begin{center}
\begin{adjustbox}{angle=0}
\begin{tabular}{|l|r|r|r|r|}\hline
\multirow{2}{0pt} &  \multicolumn{2}{c|}{\bf Gender} &  \multicolumn{2}{c|}{\bf Skin Color/Type} \\\hline\hline
                  \bf  Data set& \bf Female & \bf Male & \bf Darker & 
                  \bf Lighter \\ \hline \hline 
LFW \cite{LFW_Tech07}	         & 22.5\% & 77.4\% & 18.8\% &  81.2\%  \\ \hline
IJB-C \cite{IJB-C_ICB18}         & 37.4\% & 62.7\% & 18.0\% & 82.0\% \\ \hline
Pubfig \cite{Pubfig_PAMI11}      & 50.8\% & 49.2\% & 18.0\% & 82.0\% \\ \hline
CelebA \cite{CelebA_ICCV15}      & 58.1\% & 42.0\% & 14.2\% & 85.8\% \\ \hline
UTKface \cite{UTKface_CVPR17}    & 47.8\% & 52.2\% & 35.6\% & 64.4\%  \\ \hline
AgeDB \cite{AgeDB_CVPRW17}       & 40.6\% & 59.5\% & 5.4\% & 94.6\% \\ \hline
PPB \cite{gendershades_FAT18}   & 44.6\% & 55.4\% & 46.4\% &	53.6\% \\ \hline
IMDB-Face \cite{imdbface_ECCV18} & 45.0\% & 55.0\% & 12.0\% & 88.0\% \\ \hline
\end{tabular}
\end{adjustbox}
\end{center}
\vspace{-0.5cm}
\caption{Distribution of gender and skin color/type for seven prominent face image data sets.}
\label{tab:sets_distribution_2}
\end{table}
\end{small}

Similarly, Table~\ref{tab:sets_distribution_2} shows the distribution of gender and skin color/type for eight face image data sets.  LFW is highly skewed towards male faces with $77.4\%$ corresponding to male.  Six of the eight data sets have more male faces.  A similar skew is seen with skin color/type when grouped coarsely into darker and lighter groups.  Note that different methods were used for characterizing skin color/type in Table~\ref{tab:sets_distribution_2}, and the meaning of darker and lighter is not the same across these eight data sets.   For all but two data sets the distribution shows $>80\%$ faces that are lighter.  AgeDb is the most heavily skewed, with $94.6\%$ faces that are lighter.   The Pilot Parliaments Benchmark (PPB) data set was designed to be balanced for gender and skin type, where a board certified dermatologist provided the ground-truth labels using the Fitzpatrick six-point system~\cite{gendershades_FAT18,Fitzpatrick88}.  However, the age distribution in PPB is skewed, having been built from official photos of members of parliaments, all adults, from six countries. Face recognition systems developed from skewed training data are bound to produce biased models.  This mismatch has been evidenced in the significant drop in performance for different groupings of faces~\cite{MITReview,NTY_FR,Unmasking_AI}.  A published study showed that gender estimation from face images is biased against dark-skinned females over white-skinned males~\cite{gendershades_FAT18,raji_aeis19}.  Such biases may have serious impacts in practice.  Yet much of the prior research on face recognition does not take these issues under consideration, having focused strongly on driving up accuracy on narrow data sets.  Note also that the gender categorization in Table~\ref{tab:sets_distribution_2}, as in much of the prior work, uses a binary system for gender classification that corresponds to biological sex -- male and female.  However, different interpretations of gender in practice can include biological gender, psychological gender and social gender roles.  As with race and ethnicity, over-simplification of gender by imposing an incomplete system of categorization can result in face recognition technologies that do not work fairly for all of us.  Some recent efforts, such as InclusiveFaceNet~\cite{inclusivefacenet}, show that imperfect categorization of race and gender can help with face attribute recognition.  However, we expect that more nuanced treatment of race, ethnicity and gender is important towards improving diversity in face data sets.

\subsection{Bias and Fairness} 

The study of bias and fairness has recently gained broad interest in computer vision and machine learning \cite{lohia2018bias,zhao2017men,Hendricks2018}.  
Torralba and Efros \cite{Torralba_CVPR11} presented an evaluation of metrics related to bias and framed bias in visual classification as a domain transfer problem. Tommasi et al. \cite{tommasi2017deeper} and Hoffman et al. \cite{url_arxiv_Hoffman13} conducted a similar evaluation with deep features, showing that data set bias can be reduced but not eliminated.  Khosla et al. \cite{Khosla_ECCV12} proposed a method that learns bias vectors associated with individual data sets, as well as weights common across data sets, which are learned by undoing unwanted bias from each data set. Hardt et al.~\cite{Hardt_NIPS16} proposed a framework for fairness called equalized odds, also referred to as disparate mistreatment \cite{Zafar_WWW17}, where the goal is to predict a true outcome based on labeled training data, while ensuring it is `non-discriminatory' with respect to a chosen protected attribute. More recently, Burns et al.~\cite{Hendricks2018} addressed bias in image captioning, proposing a model that ensures equal gender probability when gender evidence is occluded in a scene, and otherwise predicts gender from relevant information when present.  The problem of gender-neutral smile classification was addressed in~\cite{inclusivefacenet}.  Bias in face detection for skin tone, pose and expression was studied in~\cite{url_arxiv_McDuff18}. Buolamwini et al. \cite{gendershades_FAT18} proposed an approach to evaluate unwanted bias in face recognition and data sets with respect to phenotypic subgroups and introduced the PPB data set balanced by gender and skin type.

\subsection{Face Data Sets} 

As described above, the last decade has seen an ever-growing collection of face recognition data sets.  Table~\ref{tab:face_db} summarizes many of the prominent face image data sets used for evaluating face recognition technology. Returning to Labeled Faces in the Wild (LFW) \cite {LFW_Tech07}, it presented considerable technical challenges upon its release in 2007, whereas nearly perfect results are being attained today. Several data sets such as IJBC \cite{maze2018iarpa}, UMD \cite{UMDFaces_ARXIV16} and VGGFace \cite{VGG_BMVC15, VGG2_FG18} provide a larger set of face images with a wider range of pose and lighting variations. Other large-scale face recognition data sets include MegaFace \cite{Megaface_CVPR16}, MS-Celeb \cite{MSCELEB1M_ECCV16} and CASIA \cite{CASIA_ARXIV14}. Many other data sets have been proposed for different facial analysis tasks, such as age and gender classification \cite{CACD_TMM15, Levi_CVPRW15,Escalera_IJCNN17,imdbface_ECCV18, Adience_TIFS14,age14}, facial expression analysis \cite{Escalera_IJCNN17}, memorability \cite{Bainbridge_JEPG2013}, attributes \cite{Pubfig_PAMI11} and aging \cite{FGNET_IET_BIOMETRICS16}. Unlike the prior data sets, which focus on robustness under variations of pose, lighting, occlusion, and scale, the $DiF$ data set is aimed understanding diversity with respect to intrinsic facial features. 

\begin{table}[!htbp]
\begin{center}
\begin{adjustbox}{angle=90}
\begin{small}
\begin{tabular}{|l|r|c|c|c|c|c|c|c|c|c|c|c|}\hline
 \rule{0pt}{37pt} {Data set} & \parbox[c]{0.5in}{Size\\\#Images} &  
  \parbox[m]{0.35in}{Iden-tity\\} & 
 \parbox[m]{0.45in}{Cranio\\-facial\\Dist.} & 
 \parbox[m]{0.45in}{Cranio\\-facial\\Areas} &
 \parbox[c]{0.45in}{Cranio\\-facial\\Ratios} &
\parbox[c]{0.45in}{Facial\\Sym-\\metry} & 
\parbox[c]{0.45in}{Facial\\Con-\\trast} & 
\parbox[c]{0.4in}{Skin\\Color/\\Type} & 
\parbox[c]{0.3in}{Age\\} & 
\parbox[c]{0.3in}{Gen-der\\} & 
\parbox[c]{0.45in}{Subj.\\Anno-tation\\} & 
\parbox[c]{0.45in}{Pose} \\ \hline \hline
MS-Celeb-1M \cite{MSCELEB1M_ECCV16}	& 8.2M & y & . & . & . & . & . & . & . & . & . & . \\ \hline
Megaface \cite{Megaface_CVPR16} &	4.7M   & y & . & . & . & . & . & . & . & y & y & . \\ \hline
VGG2 \cite{VGG2_FG18} & 3.3M              & y & . & . & . & . & . & . & . & y & . & . \\ \hline
VGG \cite{VGG_BMVC15} 	& 2.6M             & y & . & . & . & . & . & . & . & . & . & . \\ \hline
IMDB-Face \cite{imdbface_ECCV18} &	1.7M   & y & . & . & . & . & . & . & . & . & . & . \\ \hline
IMDB-Wiki \cite{age_image18} &523,051 & y & . & . & . & . & . & . & y & y & . & . \\ \hline
Casia-Webface \cite{CASIA_ARXIV14}&	494,414& y & . & . & . & . & . & . & . & y & y & . \\ \hline
UMDFaces \cite{UMDFaces_ARXIV16} & 367,920 & y & . & . & . & . & . & . & y & y & . & . \\ \hline
CelebA \cite{CelebA_ICCV15}  & 202,599	   & . & . & . & . & . & . & . & . & y & y & y \\ \hline
CACD \cite{CACD_TMM15} &	163,446        & y & . & . & . & . & . & . & y & y & y & . \\ \hline
IJB-C \cite{IJB-C_ICB18} &	141,332	       & y & . & . & . & . & . & y & y & y & y & y \\ \hline
FaceScrub \cite{FaceScrub_ICIP14} & 105,830& . & . & . & . & . & . & . & y & y & y & . \\ \hline
IJB-B \cite{IJB-B_CVPRW17} & 68,195        & y & . & . & . & . & . & y & y & y & y & y \\ \hline
Pubfig \cite{Pubfig_PAMI11} & 58,797       & y & . & . & . & . & . & . & . & y & y & y \\ \hline
Morph \cite{morph_FG06}&	55,134         & y & . & . & . & . & . & y & y & y & y & . \\ \hline
Adience \cite{Adience_TIFS14} &26,580      & . & . & . & . & . & . & . & y & y & y & . \\ \hline
UTKface \cite{UTKface_CVPR17} & 24,108     & . & . & . & . & . & . & . & y & y & y & . \\ \hline
AgeDB \cite{AgeDB_CVPRW17}  & 16,488       & . & . & . & . & . & . & . & y & y & y & . \\ \hline
LFW(A) \cite{LFW_Tech07}	& 13,233       & y & . & . & . & . & . & . & y & y & y & . \\ \hline
LFW+ \cite{lfwplus_TPAMI17}	& 	15,699     & y & . & . & . & . & . & . & . & y & y & . \\ \hline
IJB-A \cite{IJB-A_CVPR15} & 5,712          & y & . & . & . & . & . & . & y & y & y & y \\ \hline
PPB \cite{gendershades_FAT18} &1,270       & . & . & . & . & . & . & y & . & y & y & . \\ \hline
FGNet \cite{FGNET_IET_BIOMETRICS16} &1,002 & . & . & . & . & . & . & y & y & y & y & . \\ \hline
\hline
{Diversity in Faces} & {0.97M} & {.} & {y}& {y}& {y} & {y} & {y} & {y} & {y} & {y} & {y} &{y}\\ \hline
\end{tabular}
\end{small}
\end{adjustbox}
\end{center}
\vspace{-0.5cm}
\caption{Summary of prominent face image data sets. The $DiF$ data set provides the most comprehensive set of annotations of intrinsic facial features, which include craniofacial distances, areas and ratios, facial symmetry and contrast, skin color/type, age and gender predictions, subjective annotations (age, gender), and pose.}
\label{tab:face_db}
\end{table}


\section{$DiF$ Data Set Construction}  \label{sec:data set}

Given the above issues, we were motivated to create the $DiF$ data set to obtain a scientific and computationally practical basis for ensuring fairness and accuracy in face recognition.  At one extreme the challenge of diversity could be solved by building a data set comprised from the face of every person in the world.  However, this would not be practical or even possible, let alone the significant privacy concerns.  For one, our facial appearances are constantly changing due to ageing, among other factors.  At best this would give a solution for a point in time.  Rather, a solution needs to come from obtaining or generating a representative sample of faces with sufficient coverage and balance.  That, however, is also not a simple task.  There are many challenging questions: what does coverage mean computationally?  How should balance be measured?  Are age, gender and skin color sufficient?  What about other highly personal attributes that are part of our identity, such as race, ethnicity, culture, geography, or visible forms of self-expression that are reflected in our faces in a myriad of ways?  We realized very quickly that until these questions were answered we could not construct a complete and balanced data set of face images.

We formulated a new approach that would help answer these questions.  We designed the $DiF$ data set to provide a scientific foundation for research into facial diversity.  We reviewed the scientific literature on face analysis and studied prior work in fields as diverse as psychology, sociology, dermatology, cosmetology and facial surgery.  We concluded that no single facial feature or combination of commonly used classifications -- such as age, gender and skin color -- would suffice.  Therefore, we formulated a novel multi-modal approach that incorporates a diversity of face analysis methods.  Based on study of the large body of prior work, we chose to implement a solid starter-set of ten facial coding schemes.  The criteria for selecting these coding schemes included several important considerations: (1) strong scientific basis as evidenced by highly cited prior work, (2) extracting the coding scheme was computationally feasible, (3) the coding scheme produced continuous valued dimensions that could feed subsequent analysis, as opposed to generating only categorical values or labels, and (4) the coding scheme would allow for human interpretation to help with our understanding.

We chose YFCC-100M~\cite{yahoo100m_CACM16} to be the source for the sample of images.  There were a number of important reasons for this.  Ideally, we would be able to automatically obtain any large sample of images from any source meeting any characteristics of diversity we desire.  However, practical considerations prevent this, including the fact that various copyright laws and privacy regulations must be respected.  YFCC-100M is one of the largest image collections, consisting of more than $100$ millions photos.  It was populated by users of the Flickr photo service.  There is a large diversity in these photos overall, where people and faces appear in an enormous number of ways.  Also, importantly, a large portion of the photos have Creative Commons license.  The downside of using YFCC-100M is that there is skew in the Flickr user community that contributed the photos.  We cannot rely on the set of users or their photos to be inherently diverse.  A consequence of this is that the set of images used in the $DiF$ is not completely balanced on its own.  However, it still provides the desired basis for studying methods for characterizing facial diversity. 

\begin{figure}
\begin{center}
\begin{tabular}{ccc}
   \includegraphics[width=0.3\linewidth]{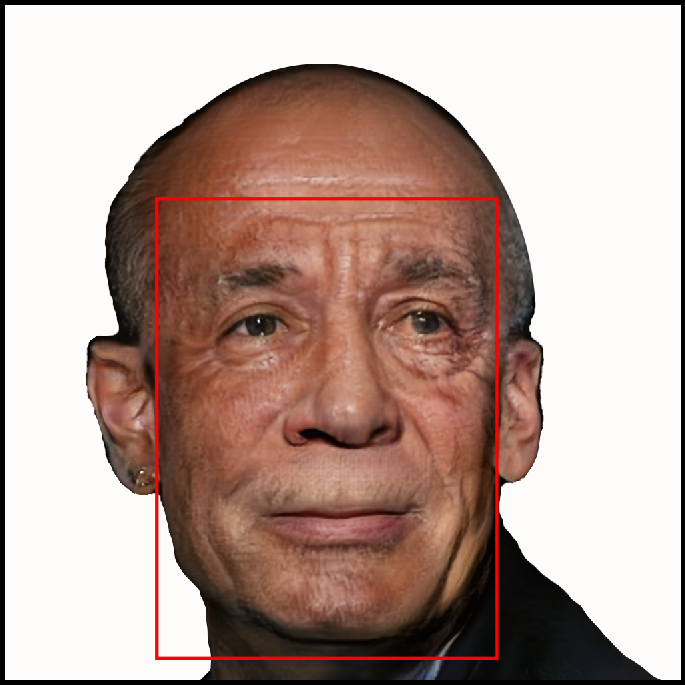} &  
   \includegraphics[width=0.3\linewidth]{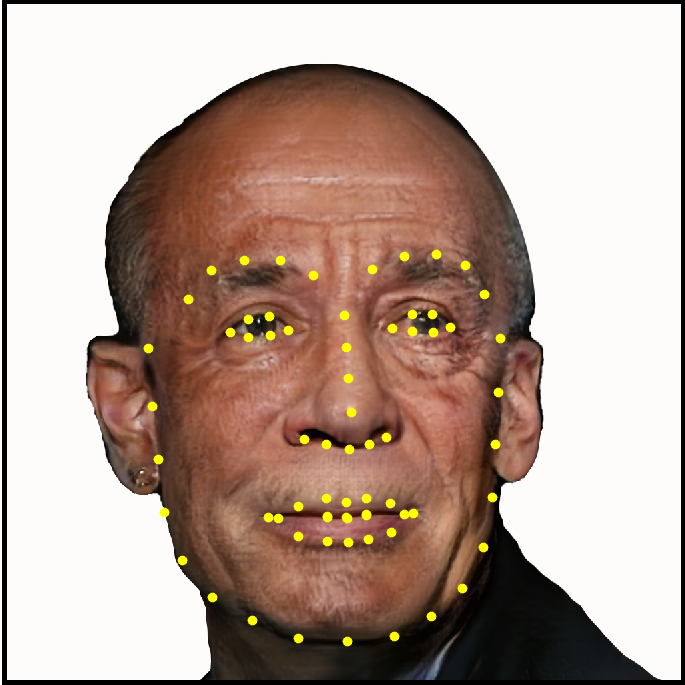} & \includegraphics[width=0.3\linewidth]{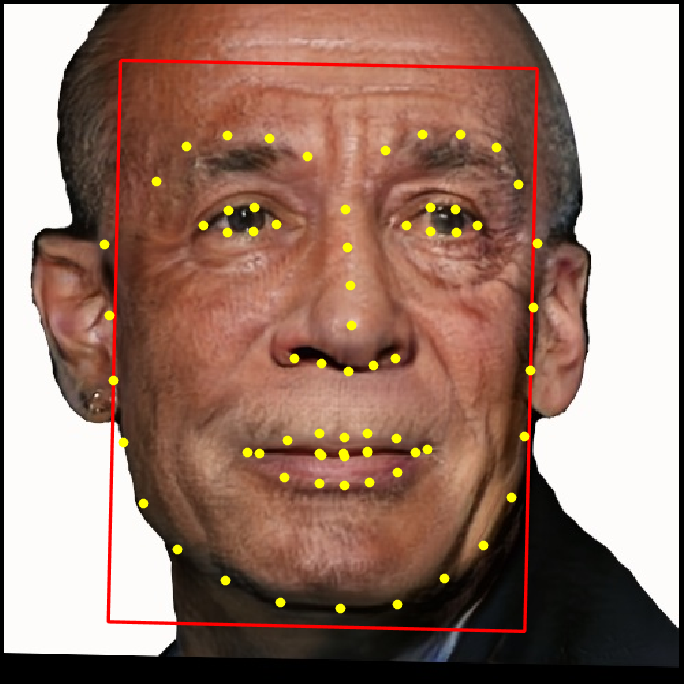} \\
   (a) face detection (size)& (b) face keypoints (pose and iod) &(c) face rectification \\
\end{tabular}
\end{center}
   \caption{Each candidate photo from YFCC-100M was processed by first detecting the depicted faces with a Convolutional Neural Network (CNN) using the Faster-RCNN based object detector~\cite{fasterrcnn_NIPS15}.  Then each detected face as in (a) was processed using DLIB~\cite{DLIB09} to extract pose and landmark points as shown in (b) and subsequently assessed based on the width and height of the face region. Faces with region size less than 50x50 or inter-ocular distance of less than $30$ pixels were discarded.  Faces with non-frontal pose, or anything beyond being slightly tilted to the left or the right, were also discarded. Finally, an affine transformation was performed using center points of both eyes, and the face was rectified as shown in (c).}
\label{fig:overall}
\end{figure}

\subsection{Data Selection}

While the YFCC-100M photo data set is large, not all images could be considered.  Naturally, we excluded photos that did not contain a face.  We also excluded black and white and grayscale photos and those with significant blur.  Although face recognition needs to be robust for non-color photos, we deferred incorporating these images in the initial $DiF$ data set in order to focus on intrinsic facial variation rather than image variation due to color processing.

\subsection{Pre-processing Pipeline}

The YFCC-100M data set gives a set of URLs that point to the Flickr web page for each of the photos.   The first step we took was to check whether the URL was still active.  If so, we then checked the license.  We proceeded with the download only if the license type was Creative Commons. Once we retrieved the photo, we processed it using face detection to find all depicted faces.  For the face detection step, we used a Convolutional Neural Network (CNN) object detector trained for faces based on Faster-RCNN \cite{fasterrcnn_NIPS15}. For each detected face, we then extracted both pose and $68$ face key-points using the open source DLIB toolkit~\cite{DLIB09}. If there was any failure in the image processing steps, we excluded the face from further consideration.  We also removed faces of size less than $50 \times 50$ pixels or with inter-ocular distance of less than $30$ pixels.  We removed faces with substantial non-frontal pose.  The overall process is shown in Figure \ref{fig:overall}.

Finally, we generated two instances of each face.  One is a rectified instance whereby the center points of each eye are fixed to a specific location in the overall image.  The second crops an expanded region surrounding each face to give 50\% additional spatial context.  This overall process filtered the $100$ million YFCC-100M photos down to approximately one million mostly frontal faces with adequate size.  The surviving face images were the ones used for the $DiF$ data set.  Note that the overall process of sampling YFCC-100M used only factors described above including color, size, quality and pose.  We did not bias the sampling towards intrinsic facial characteristics or by using metadata associated with each photo, such as a geo-tag, date, labels or Flickr user name.  In this sense, the $DiF$ data distribution is expected to closely follow the overall distribution of the YFCC-100M photos.  In future efforts to grow the $DiF$ data set, we may relax some of the constraints based on size, pose and quality, or we may bias the sampling based on other properties.  However, one million publicly available face images provides a good start.  Given this compiled set of faces, we next process each one by extracting the ten facial coding schemes.


\section{Facial Coding Scheme Implementation}  \label{sec:codingschemes}

In this Section, we describe the implementation of the ten facial coding schemes and the process of extracting them from the $DiF$ face images.  The advantage of using ten coding schemes is that it gives a diversity of methods and allows us to compare statistical measures for facial diversity.  As described above, the ten schemes have been selected based on their strong scientific basis, computational feasibility, numerical representation and interpretability.  Overall the chosen ten coding schemes capture multiple modalities of facial features, which includes craniofacial distances, areas and ratios, facial symmetry and contrast, skin color, age and gender predictions, subjective annotations, and pose and resolution.  Three of the $DiF$ facial coding schemes are based on craniofacial features.  As prior work has pointed out, skin color alone is not a strong predictor of race, and other features such as facial proportions are important~\cite{learning_race14,race_variation79,anthro_woman01,facial_contrast17}.  Face morphology is also relevant for attributes such as age and gender~\cite{face_age_modeling06}.  We incorporated multiple facial coding schemes aimed at capturing facial morphology using craniofacial features~\cite{anthropometry_book94,anthropmetry05,face_age_modeling06}.  The basis of craniofacial science is the measurement of the face in terms of distances, sizes and ratios between specific points such as the tip of the nose, corner of the eyes, lips, chin, and so on.  Many of these measures can be reliably estimated from photos of frontal faces using $47$ landmark points of the head and face~\cite{anthropometry_book94}.  To provide the basis for the three craniofacial feature coding schemes used in $DiF$, we built on the subset of $19$ facial landmarks listed in Table \ref{tab:landmarks}.  For brevity we adopt the abbreviations from~\cite{anthropometry_book94} when referring to these facial landmark points instead of using the full anatomical terms.

\begin{table}[ht]
\centering
\begin{tabular}{|lc|lc|}
    \hline 
   \bf Anatomical term & \bf Abbreviation  & \bf Anatomical term & \bf Abbreviation\\
     \hline  \hline
    {\it tragion} & $tn$ & {\it subalare} &$sbal$ \\ \hline
    {\it orbitale} &  $or$ & {\it subnasale} & $sn$\\ \hline
    {\it palpebrale superius} & $ps$ & {\it crista philtre} &$cph$\\ \hline
    {\it palpebrale inferius} & $pi$ & {\it labiale superius} &$ls$ \\ \hline
    {\it endocanthion} & $en$ & {\it stornion} & $sto$\\ \hline
    {\it exocanthion} & $ex$ & {\it labiale inferius} & $li$\\ \hline
    {\it nasion} & $n$  & {\it chelion}& $ch$  \\ \hline
    {\it pronasale} & $c'$ & {\it gonion} & $go$\\ \hline
    {\it zygion} & $zy$ & {\it gnathion} & $gn$ \\ \hline
    {\it alare} &$al$ & & \\ \hline
    \end{tabular}
    \caption{Anatomical terms and corresponding abbreviations (as in \cite{anthropometry_book94}) for the set of facial landmarks employed to compute the craniofacial measurements for facial coding schemes 1--3.}
    \label{tab:landmarks}
\end{table}

\begin{figure}[ht]
\begin{center}
\includegraphics[width=0.95\linewidth]{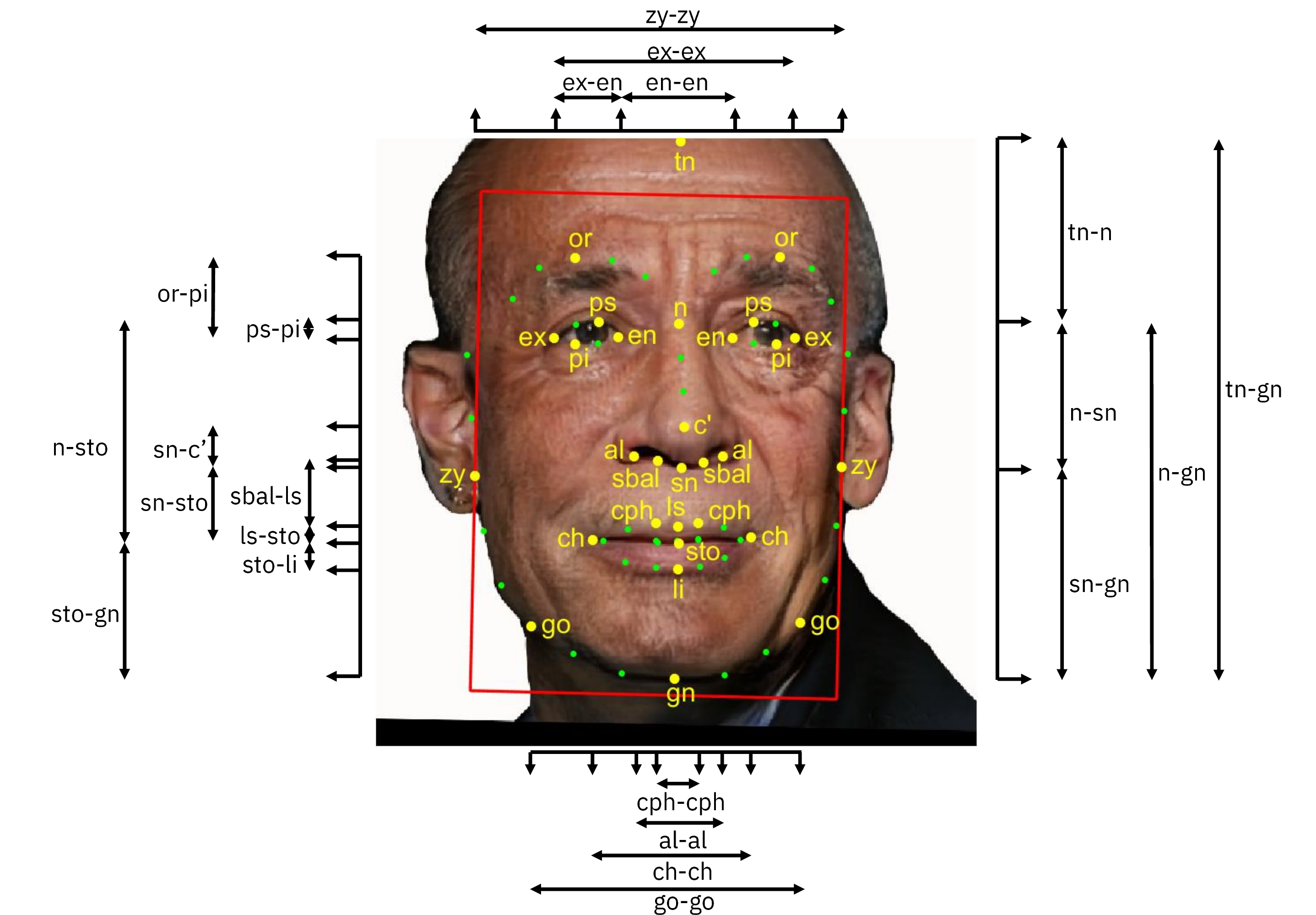}
\end{center}
\caption{We used the $68$ key-points extracted using DLIB from each face (small dots) to localize $19$ facial landmarks (large dots, labeled), out of the 47 introduced in~\cite{anthropometry_book94}. Those 19 landmarks were employed as the basis for extraction of the craniofacial measures for coding schemes 1--3. }
\label{fig:cranio}
\end{figure}

In order to extract the $19$ facial landmark points, we leveraged standard DLIB facial key-point extraction tools that provide a set $68$ key-points for each face.  As shown in Figure~\ref{fig:cranio}, we mapped the $68$ DLIB key-points to the $19$ facial landmarks~\cite{anthropometry_book94}.  These $19$ landmarks were used for extracting the craniofacial features.  Note that for illustrative purposes, the example face used in Figure~\ref{fig:cranio} was adopted from~\cite{progressiveGAN_ICLR18} and was generated synthetically using a progressive Generative Adversarial Network (GAN) model.  The face does not correspond to a known individual person.  However, the image is subject to license terms as per~\cite{progressiveGAN_ICLR18}.  In order to incorporate a diversity of approaches, we implemented three facial coding schemes for craniofacial features.  The first, coding scheme 1, provides a set of craniofacial distance measures from~\cite{anthropometry_book94}.  The second, coding scheme 2, provides an expanded set of craniofacial areas from~\cite{anthropmetry05}.  The third, coding scheme 3, provides a set of craniofacial ratios from~\cite{face_age_modeling06}.


\subsection{Coding Scheme 1: Craniofacial Distances} \label{sec:cf1}

The first coding scheme for craniofacial distances has been adopted from~\cite{anthropometry_book94}. It comprises eight measures which characterize all the vertical distances between elements in a face: the top of the forehead, the eyes, the nose, the mouth and the chin.  In referring to the implementation of the coding scheme, we use the abbreviations from Table~\ref{tab:landmarks}.  We note that two required points, $tn$ and $sto$, were not part of the set of 68 DLIB key-points. As such, we had to derive them in the following manner: $tn$ was computed as the topmost point vertically above $n$ in the rectified facial image, and $sto$ was computed from the vertical average of $ls$ and $li$.  The eight dimensions of craniofacial distances are summarized in Table~\ref{tab:cranio1}.

\begin{table}[ht]
\centering
\begin{tabular}{|l|c|}
    \hline
   \bf Craniofacial distance    & \bf Measure  \\ \hline\hline
    {\it intercanthal} face height & $n-sto$ \\ \hline
    eye fissure height (left and right) & $ps-pi$ \\ \hline
    orbit and brow height (left and right) & $or-pi$ \\ \hline
    {\it columella} length& $sn-c'$ \\ \hline
    upper lip height& $sn-sto$ \\ \hline
    lower {\it vermilion} height& $sto-li$ \\ \hline
    {\it philtrum} width & $cph-cph$ \\ \hline
    lateral upper lip heights  (left and right)& $sbal-ls'$\\ \hline
    \end{tabular}
    \vspace{-0.1cm}
    \caption{Coding scheme 1 is made up eight craniofacial measures corresponding to different vertical distances in the face~\cite{anthropometry_book94}.}
    \label{tab:cranio1}
\end{table}


\subsection{Coding Scheme 2: Craniofacial Areas} \label{sec:cf2}

 The second coding scheme is adopted from a later development from Farkas et al.~\cite{anthropmetry05}.  It comprises measures corresponding to different areas of the cranium.  Similar to the craniofacial distances, the extraction of craniofacial areas relied on the mapped DLIB key-points to the corresponding facial landmarks.  Table~\ref{tab:cranio2} summarizes the twelve dimensions of the craniofacial area features. 
 
\begin{table}[ht]
\centering
\begin{tabular}{|l|c|}
    \hline 
 \bf  Craniofacial area    & \bf Measure\\\hline\hline
    Head height& $tn-n$   \\ \hline 
    Face height & $tn-gn$  \\ \hline
    Face height& $n-gn$  \\ \hline
    Face height& $sn-gn$ \\ \hline
    Face width& $zy-zy$ \\ \hline
    Face  width& $go-go$ \\ \hline
    {\it Orbits intercanthal} width & $en-en$ \\ \hline
    {\it Orbits fissure} length (left and right)& $en-ex$  \\ \hline
    {\it Orbits biocular} width& $ex-ex$ \\ \hline
    Nose height& $n-sn$ \\ \hline
    Nose width& $al-al$ \\ \hline
    {\it Labio-oral} region & $ch-ch$ \\ \hline
    \end{tabular}
    \vspace{-0.1cm}
    \caption{Coding scheme 2 is made up of twelve craniofacial measures that correspond to different areas of the face~\cite{anthropmetry05}.}
    \label{tab:cranio2}
\end{table}


\subsection{Coding Scheme 3: Craniofacial Ratios} \label{sec:cf3}

 The third coding scheme comprises measures corresponding to different ratios of the face. These features were used to estimate age progression from faces in the age groups of 0 to 18 in~\cite{face_age_modeling06}. Similar to the above features, the craniofacial ratios used the mapped DLIB key-points as facial landmarks.  Table~\ref{tab:cranio3} summarizes the eight dimensions of the craniofacial ratio features. 

\begin{table}[ht]
\centering
\begin{tabular}{|l|c|}
    \hline 
   \bf Craniofacial ratio    & \bf Measure\\\hline\hline
    Facial index& $(n-gn)/(zy-zy)$ \\ \hline
    {\it Mandibular} index& $(sto-gn)/(go-go)$\\ \hline
    {\it Intercanthal} index& $(en-en)/(ex-ex)$\\ \hline
    {\it Orbital} width index (left and right)& $(ex-en)/(en-en)$\\ \hline
    Eye fissure index (left and right)& $(ps-pi)/(ex-en)$\\ \hline
    Nasal index& $(al-al)/(n-sn)$\\ \hline
    {\it Vermilion} height index& $(ls-sto)/(sto-li)$\\ \hline
    Mouth-face width index& $(ch-ch)/(zy-zy)$\\ \hline
    \end{tabular}
    \vspace{-0.1cm}
    \caption{Coding scheme 3 is made up of eight craniofacial measures that correspond to different ratios of the face~\cite{anthropmetry05}.}
    \label{tab:cranio3}
\end{table}

\subsection{Coding Scheme 4: Facial Symmetry}

\begin{figure}[h]
\begin{center}
\begin{tabular}{ccc}
   \includegraphics[width=0.25\linewidth]{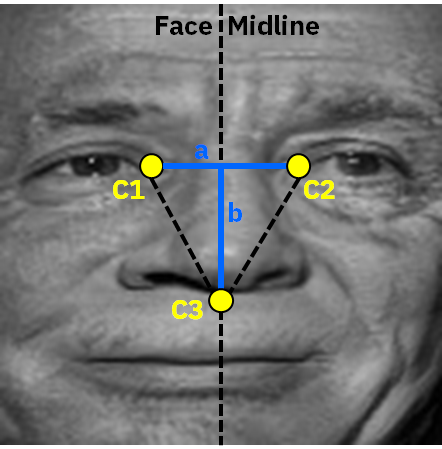}  &  
   \includegraphics[width=0.25\linewidth]{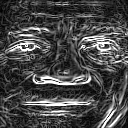}  & 
     \includegraphics[width=0.25\linewidth]{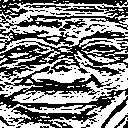} \\
     (a) & (b) & (c)
\end{tabular}
\end{center}
\vspace{-0.5cm}
\caption{Process for extracting facial symmetry measures for coding scheme 4, starting with (a) rectified face showing face mid-line and reference points for inner {\it canthus} (C1 and C2) and {\it philtrum} (C3) and line segmented connecting them (point $a$ for C1-C2 and point $b$ connecting C3 to the midpoint of point $a$).  Additionally, a Sobel filter is used to extract (b) edge magnitude and (c) orientation to derive the measure for edge orientation similarity.}
\label{fig:symmetry2}
\end{figure}

Facial symmetry has been found in  psychology and anthropology studies to be correlated with  subjective and objective traits including expression variation~\cite{symmetryFG02} and attractiveness~\cite{facial_attractiveness11}.  We adopted facial symmetry for coding scheme 4, given its intrinsic nature.  To represent the symmetry of each face we computed two measures, following the work of Liu et al.~\cite{symmetryFG02}.  We processed each face as shown in Figure~\ref{fig:symmetry2}.  We used three of the DLIB key-points detected in the face image to spatially normalize and rectify it to the following locations: the inner {\it canthus} of each eye (C1 and C2) to reference locations $C1=(40,48)$, $C2=(88,48)$ and the {\it philtrum} C3 was mapped to $C3=(64,84)$.  Next, the face mid-line (point $b$ in Figure \ref{fig:symmetry2}(a)) was computed as the line passing through the mid-point of the line segment connecting $C1-C2$ (point $a$ in Figure \ref{fig:symmetry2}(a)) and the {\it philtrum} C3.

We point out that although a face image is spatially transformed during rectification, facial symmetry with respect to the face mid-line is preserved according to the topological properties of the affine transformation~\cite{topology77}.  Each image is then cropped to 128x128 pixels to create a squared image with the face mid-line centered vertically.  Next we convert the spatially transformed image to grayscale to measure intensity.  Each point $(x,y)$ on this normalized face intensity image $I$ on the left of the face mid-line has a unique corresponding horizontally mirrored point on the other side of the face image $I'(x,y)$ (right of the mid-line).  We also extract edges in this image $I$ to produce $I_e$ using a Sobel filter.  Finally, we compute two facial symmetry measures based on density difference $DD(x,y)$ and edge orientation similarity $EOS(x,y)$ as follows:  for each pixel $(x,y)$ in the left 128x64 part ($I$ and $I_e$) and the corresponding 128x64 right part ($I'$ and $I'_e$) are computed as summarized in Table~\ref{tab:symmetry}, where $\phi(I_e(x,y),I'_e(x,y))$ is the angle between the two edge orientations of images $I_e$ and $I'_e$ at pixel $(x,y)$.  We compute the average value of $DD(x,y)$ and $EOS(x,y)$ to be the two measures for facial symmetry.

\begin{table}[ht]
\centering
\begin{tabular}{|l|c|}
    \hline 
   \bf Facial symmetry    & \bf Measure\\\hline\hline
    Density difference& $DD(x,y) = I(x,y) -I'(x,y)$ \\ \hline
    Edge orientation similarity& $EOS(x,y) = cos(\phi(I_e(x,y),I'_e(x,y)))$\\ \hline
    \end{tabular}
    \vspace{-0.2cm}
    \caption{Coding scheme 4 is made up of two measures of facial symmetry~\cite{anthropmetry05}.}
    \label{tab:symmetry}
\end{table}

It is interesting to notice that the two symmetry measurements capture facial symmetry from different perspectives: density difference is affected by the left-right relative intensity variations of a face, while edge orientation similarity is affected by the zero-crossing of the intensity field. Higher values of density difference correspond to more asymmetrical faces, while the higher the values of edge orientation similarity refer to more symmetrical faces.

\subsection{Coding Scheme 5: Facial Regions Contrast}\label{sec:contrast}

Prior studies have shown that facial contrast is a cross-cultural cue for perceiving facial attributes such as age.  An analysis of full face color photographs of Chinese, Latin American and black South African women aged $20$--$80$ in~\cite{facial_contrast17} found similar changes in facial contrast with ageing across races and were comparable to changes with Caucasian faces.  This study found that high-contrast faces were judged to be younger than low-contrast faces.  The study also found that artificially increasing the aspects of facial contrast that decrease with age across diverse races makes faces look younger, independent of the ethnic origin of the face or cultural origin of the observers~\cite{facial_contrast17}.  On one hand, the age that you are is one dimension that needs to be addressed in terms of fairness and accuracy of face recognition.  However, the age that you look, considering possible artificial changes, should not change requirements for fairness and accuracy.  

\begin{figure}[hb]
\begin{center}
\includegraphics[width=0.39\linewidth]{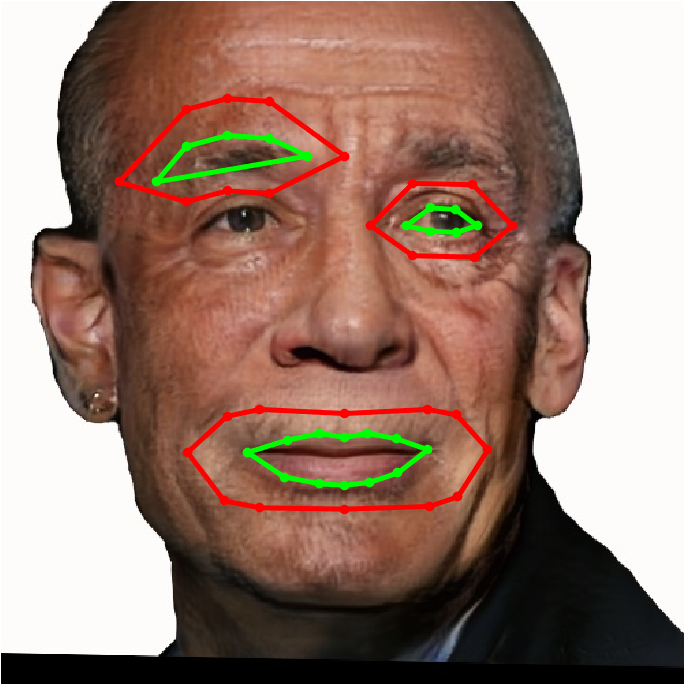}
\end{center}
\vspace{-0.6cm}
\caption{Process for extracting facial regions contrast measures for coding scheme 5.  The computation is based on the average pixel intensity differences between the outer and inner regions for the lips, eyes and eyebrows as depicted above.}
\label{fig:facial_contrast}
\end{figure}

We adopted facial regions contrast as the basis for coding scheme 5. To compute facial contrast, we measured contrast individually for each image color channel $I_{L}$, $I_{a}$, $I_{b}$, corresponding to the CIE-Lab color space, for three facial regions: lips, eyes, and eyebrows, as shown in Figure \ref{fig:facial_contrast}.  First, we defined the internal regions ringed by facial key points computed from DLIB for each of these facial parts (shown as the inner rings around lips, eyes, and eyebrows in Figure \ref{fig:facial_contrast}). Then, we expanded this region by 50\% to define an outer region around each of these facial parts (shown as the outer rings in Figure \ref{fig:facial_contrast}). The contrast is then measured as the difference between the average pixel intensities in the outer and inner regions.  This is repeated for each of the three CIE-Lab color channels.  Given the three facial regions, this gives a total of nine measures, where the contrast values for the eyes and eyebrows are based on the average of the left and right regions.  The computation is summarized in Table~\ref{tab:contrast}, where $I_{k}(x,y)$ is the pixel intensity at $(x,y)$ for CIE-Lab channel $k$ and $pt_{outer}, pt_{inner}$ correspond to the outer and inner regions around each facial part $pt$.

\begin{table}[htbp]
\centering
\begin{tabular}{|l|c|}
    \hline 
    \bf Facial region contrast    & \bf Measure\\\hline\hline
    Lips CIE-L&   $C_{L,lips} = \frac{\sum_{x,y\in lips_{outer}} I_{L}(x,y) - \sum_{x,y\in lips_{inner}} I_{L}(x,y)}{\sum_{x,y\in lips_{outer}} I_{L}(x,y) + \sum_{x,y\in lips_{inner}} I_{L}(x,y)}$\\\hline
    Lips CIE-a&   $C_{a,lips} = \frac{\sum_{x,y\in lips_{outer}} I_{a}(x,y) - \sum_{x,y\in lips_{inner}} I_{a}(x,y)}{\sum_{x,y\in lips_{outer}} I_{a}(x,y) + \sum_{x,y\in lips_{inner}} I_{a}(x,y)}$\\\hline
    Lips CIE-b&   $C_{b,lips} = \frac{\sum_{x,y\in lips_{outer}} I_{b}(x,y) - \sum_{x,y\in lips_{inner}} I_{b}(x,y)}{\sum_{x,y\in lips_{outer}} I_{b}(x,y) + \sum_{x,y\in lips_{inner}} I_{b}(x,y)}$\\\hline
    Eyes CIE-L&   $C_{L,eyes} = \frac{\sum_{x,y\in eyes_{outer}} I_{L}(x,y) - \sum_{x,y\in eyes_{inner}} I_{L}(x,y)}{\sum_{x,y\in eyes_{outer}} I_{L}(x,y) + \sum_{x,y\in eyes_{inner}} I_{L}(x,y)}$\\\hline
    Eyes CIE-a&   $C_{a,eyes} = \frac{\sum_{x,y\in eyes_{outer}} I_{a}(x,y) - \sum_{x,y\in eyes_{inner}} I_{a}(x,y)}{\sum_{x,y\in eyes_{outer}} I_{a}(x,y) + \sum_{x,y\in eyes_{inner}} I_{a}(x,y)}$\\\hline
    Eyes CIE-b&   $C_{b,eyes} = \frac{\sum_{x,y\in eyes_{outer}} I_{b}(x,y) - \sum_{x,y\in eyes_{inner}} I_{b}(x,y)}{\sum_{x,y\in eyes_{outer}} I_{b}(x,y) + \sum_{x,y\in eyes_{inner}} I_{b}(x,y)}$\\\hline
    Eyebrows CIE-L&   $C_{L,eyebrows} = \frac{\sum_{x,y\in eyebrows_{outer}} I_{L}(x,y) - \sum_{x,y\in eyebrows_{inner}} I_{L}(x,y)}{\sum_{x,y\in eyebrows_{outer}} I_{L}(x,y) + \sum_{x,y\in eyebrows_{inner}} I_{L}(x,y)}$\\\hline
    Eyebrows CIE-a&   $C_{a,eyebrows} = \frac{\sum_{x,y\in eyebrows_{outer}} I_{a}(x,y) - \sum_{x,y\in eyebrows_{inner}} I_{a}(x,y)}{\sum_{x,y\in eyebrows_{outer}} I_{a}(x,y) + \sum_{x,y\in eyebrows_{inner}} I_{a}(x,y)}$\\\hline
    Eyebrows CIE-b&   $C_{b,eyebrows} = \frac{\sum_{x,y\in eyebrows_{outer}} I_{b}(x,y) - \sum_{x,y\in eyebrows_{inner}} I_{b}(x,y)}{\sum_{x,y\in eyebrows_{outer}} I_{b}(x,y) + \sum_{x,y\in eyebrows_{inner}} I_{b}(x,y)}$\\\hline
    \end{tabular}
    \caption{Coding scheme 5 is made up of three measures of facial region contrast~\cite{facial_contrast17}.}
    \label{tab:contrast}
\end{table}

\subsection{Coding Scheme 6: Skin Color}

\begin{figure}[h]
\begin{center}
\begin{tabular}{cccc}
  \includegraphics[width=0.21\linewidth]{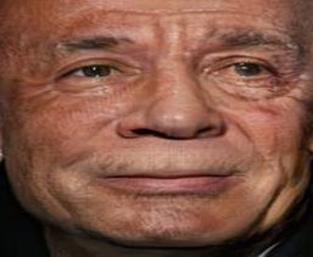}  & \includegraphics[width=0.21\linewidth]{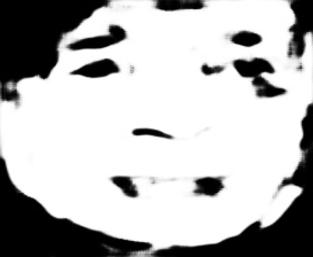} &
  \includegraphics[width=0.21\linewidth]{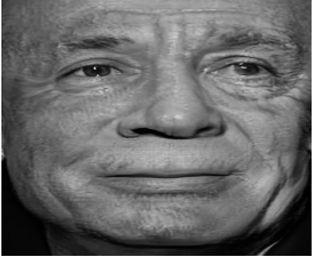} &
  \includegraphics[width=0.21\linewidth]{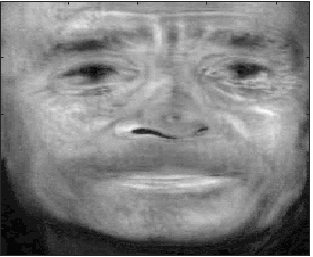} \\
  (a) & (b) & (c) & (d)  \\
    \includegraphics[width=0.21\linewidth]{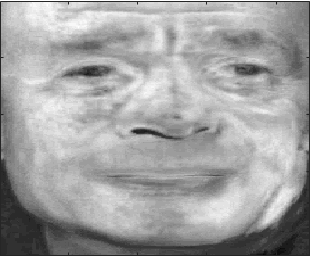}  & \includegraphics[width=0.21\linewidth]{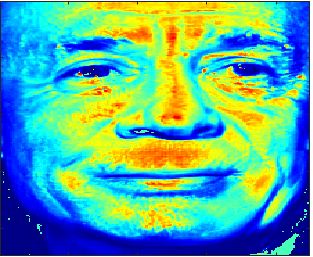} &
  \includegraphics[width=0.21\linewidth]{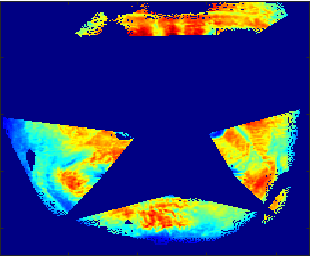} &
  \includegraphics[width=0.21\linewidth]{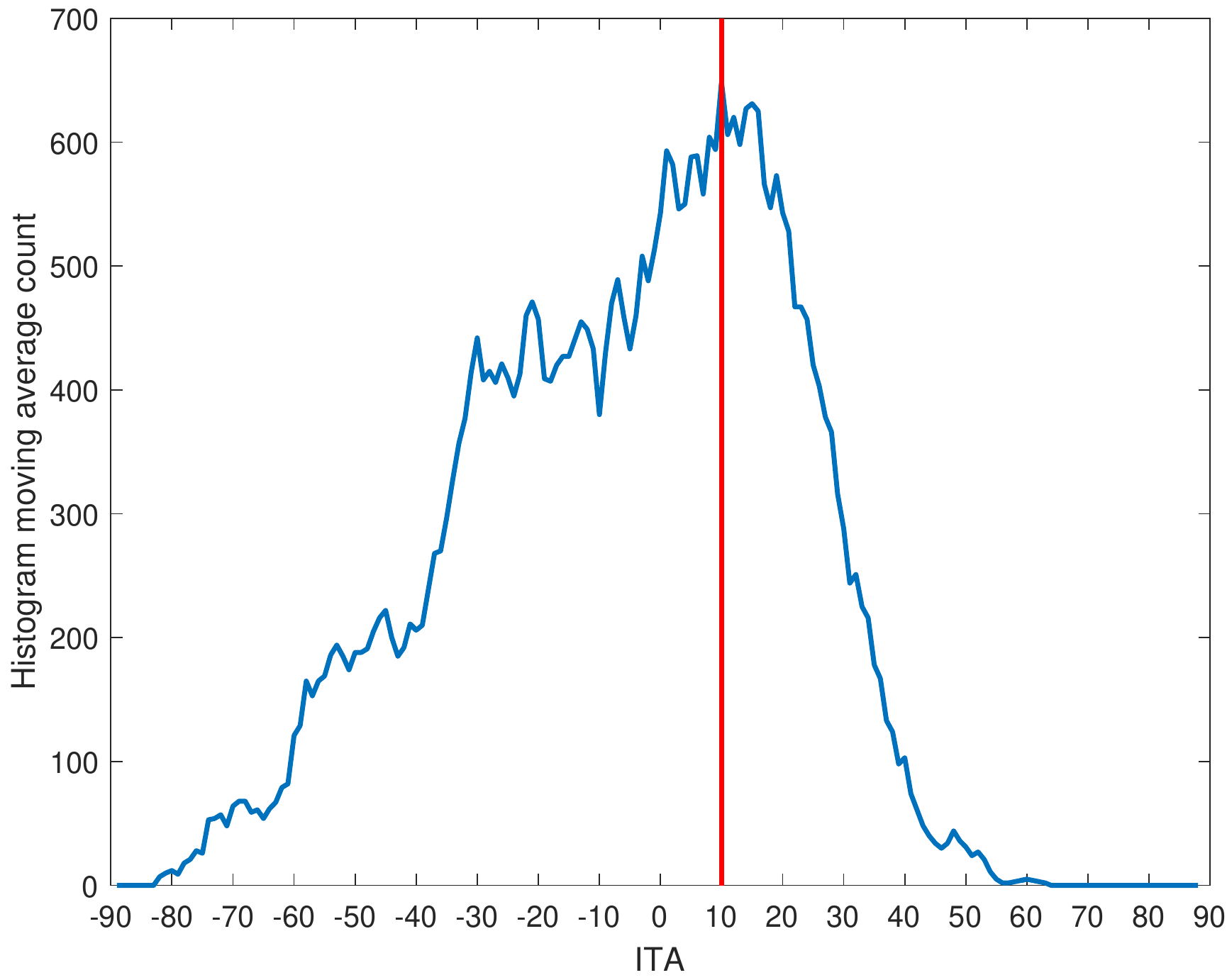} \\
  (e) & (f) & (g) & (h)  
\end{tabular}
\end{center}
\vspace{-0.5cm}
   \caption{Process for extracting skin color for coding scheme 6 based on Individual Typology Angle-based (ITA). (a) Input face (b) skin map (c) $L$ channel (d) $a$ channel (e) $b$ channel (f) ITA map (g) masked ITA map (h) ITA histogram.}
\label{fig:ita}
\end{figure}

Skin occupies a large fraction of the face.  As such, characteristics of the skin influence the appearance and perception of faces.  Prior work has studied different methods of characterizing skin based on skin color~\cite{ita_ijcs1991,takiwaki98,wang15}, skin type~\cite{ita_ijcs1991,Fitzpatrick88} and skin reflectance ~\cite{weyrich06}.  Early studies used Fitzpatrick skin type (FST) to classify sun-reactive skin types~\cite{Fitzpatrick88}, which was also adopted recently in~\cite{gendershades_FAT18}.  However, to-date, there is no universal measure for skin color, even within the dermatology field.  In a study of $556$ participants in South Africa, self-identified as either black, Indian/Asian, white, or mixed, Wilkes et al. found a high correlation between the Melanin Index (MI), which is frequently used to assign FST, with Individual Typology Angle (ITA)~\cite{wilkes15}.  Since a dermatology expert is typically needed to assign the FST, the high correlation of MI and ITA indicates that ITA may be a practical method for measuring skin color given the simplicity of computing ITA. In order to explore this further, we designed coding scheme 6 to use ITA for representing skin color~\cite{ita_ijcs1991}.  ITA has a strong advantage over Fitzpatrick in that it can be computed directly from an image.  As in~\cite{ita_ijcs1991}, we implemented ITA in the CIE-Lab space. For obvious practical reasons, we could not obtain measurements through a device directly applied on the skin of each individual, but instead converted the $RGB$ image to CIE-Lab space using standard image processing. The $L$ axis quantifies luminance, whereas $a$ quantifies absence or presence of redness, and $b$ quantifies yellowness.  Figure~\ref{fig:ita} depicts the image processing steps for extracting the coding scheme 6 for skin color.

It is important to note that that ITA is a point measurement. Hence, every pixel corresponding to skin can have an ITA measurement.  In order to generate a feature measure for the whole face, we extract ITA for pixels within a masked face region as shown in Figure \ref{fig:ita}(g).  This masked region is determined in the following steps:
\begin{enumerate}
    \item Extract the skin mask in the face (pixels corresponding to skin) using a deep neural network, as described in~\cite{noel_skin_segmentation}.
    \item Extract regions corresponding to the chin, two cheeks and forehead using the extracted 68 DLIB key-points
    \item Smooth the ITA values of each region to reduce outliers using an averaging filter
    \item Pick the peak value of each region to give its ITA score
    \item Average the values to give a single ITA score for each face
\end{enumerate}
Table~\ref{tab:skincolor} gives the formula for computing the ITA values for each pixel in the masked face region.
\begin{table}[ht]
\centering
\begin{tabular}{|l|c|}
    \hline 
    \bf Skin color    & \bf Measure\\\hline\hline
    & \\
    Individual Typology Angle (ITA)& $\frac{arctan(\frac{L-50}{b}) \times 180}{\pi}$ \\
    & \\ \hline
\end{tabular}
    \caption{Coding scheme 6 measures skin color using Individual Typology Angle (ITA)~\cite{anthropmetry05}.}
    \label{tab:skincolor}
\end{table}

\subsection{Coding Scheme 7: Age Prediction}\label{sec:age}

Age is an attribute we all possess and our faces are predictors of our age, whether it is our actual age or manipulated age appearance~\cite{facial_contrast17}.  As discussed in Section~\ref{sec:contrast}, particular facial features such as facial contrast are correlated with age.  As an alternative to designing specific feature representations for predicting age, for coding scheme 7, we adopt a Convolutional Neural Network (CNN) that is trained from face images to predict age.  We adopt the DEX model~\cite{age_image15,age_image18} that is among the highest performing on some of the known face image data sets.  The model is based on a pre-trained VGG16-face neural network for face identity that was subsequently fine-tuned on the IMDB-wiki data set \cite{age_image18} to predict age (years in the range 0-100). Since the DEX model was trained within a narrow context, it is not likely to be fair.  However, our initial use here is to get some continuous measure of age in order to study diversity.  Ultimately, it will require an iterative process of understanding diversity to make more balanced data sets and create more fair models.  In order to predict age using DEX, each face was pre-processed as in \cite{age_image15}. First, the bounding box was expanded by 40\% both horizontally and vertically, then resized to 256x256 pixels. Inferencing was then performed on the  224x224 square cropped at the center of the image.  Since softmax loss was used during the fine-tuning process, age prediction is output from the softmax layer, which is computed from $E(P) = \sum_{i=0}^{100} p_i y_i$, where $p_i \in P$ are the softmax output probabilities for each of the 101 discrete years $y_i \in Y$ corresponding to each class $i$, with $Y = \{0,...,100\}$.

\subsection{Coding Scheme 8: Gender Prediction}

Coding scheme 8 follows a similar process for gender prediction as for age prediction~\cite{age_image18}.  We used the same pre-processing steps as described in Section~\ref{sec:age} as well as the same neural network model and training pipeline. The only difference is that we use the DEX model to predict a continuous value score for gender between $0$ and $1$, and not just report a binary output.

\subsection{Coding Scheme 9: Subjective Annotation}

Coding scheme 9 aims at capturing age and gender but through subjective means rather than using a neural network-based predictive model.  For each of the $DiF$ face images, we employed the Figure Eight crowd-sourcing platform \cite{f8} to obtain subjective human-labeled annotations of gender and age.
The gender annotations used two class labels (male and female) and the age group labeling used seven classes ([0-3],[4-12],[13-19],[20-30],[31-45],[46-60],[61-]), as well as a continuous age value to be consistent with the automatic prediction labels.  
For each face, input was taken from three independent annotators.  A weighted voting scheme was used to aggregate the labels, where the vote of each annotator was weighted according to their performance on a set of ``gold standard'' faces for which the ground truth was known.

\subsection{Coding Scheme 10: Pose and Resolution}

The final coding scheme 10 provides information about pose and resolution.  Although pose can only loosely be considered an intrinsic facial attribute, how we present our faces to cameras should not affect performance.  We include resolution as it gives useful information to correlate with the other coding scheme features to provide further insight.  In order to extract pose, we use the DLIB toolkit to compute a pose score of 0-4.  Here, the values correspond as follows: 0-frontal, 1-rotated left, 2-rotated right, 3-frontal but tilted left, 4-frontal but tilted right. Along with pose, resolution is determined from the size of the bounding box of each face and inter-ocular distance (IOD), which is the distance between the center points of each eye.
 
\section{Statistical Analysis} \label{sec:analysis}
\begin{figure}[hb]
\begin{center}
\begin{tabular}{cccc}
\includegraphics[width=0.23\linewidth]{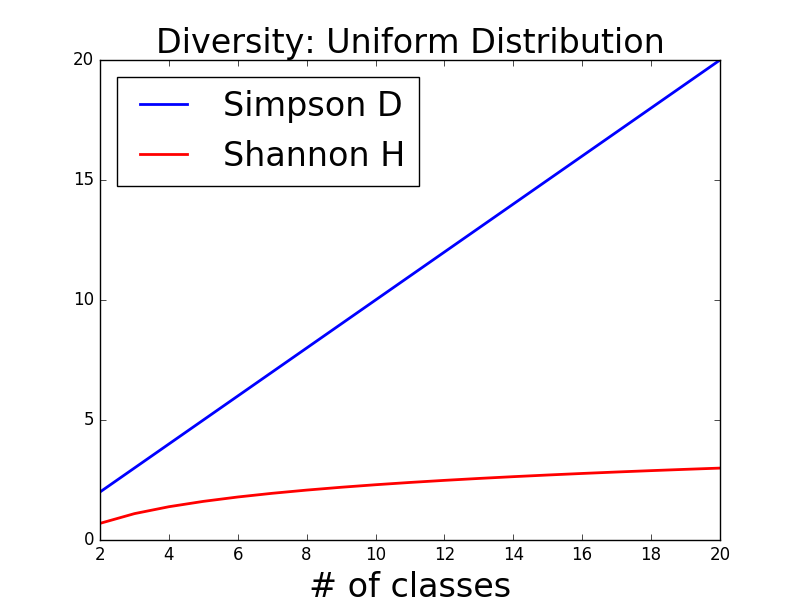} &
\includegraphics[width=0.23\linewidth]{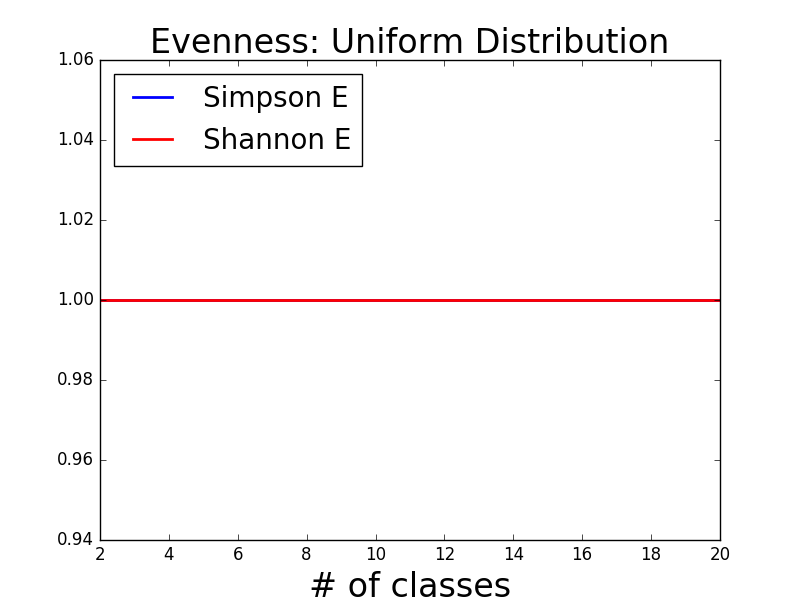} & 
\includegraphics[width=0.23\linewidth]{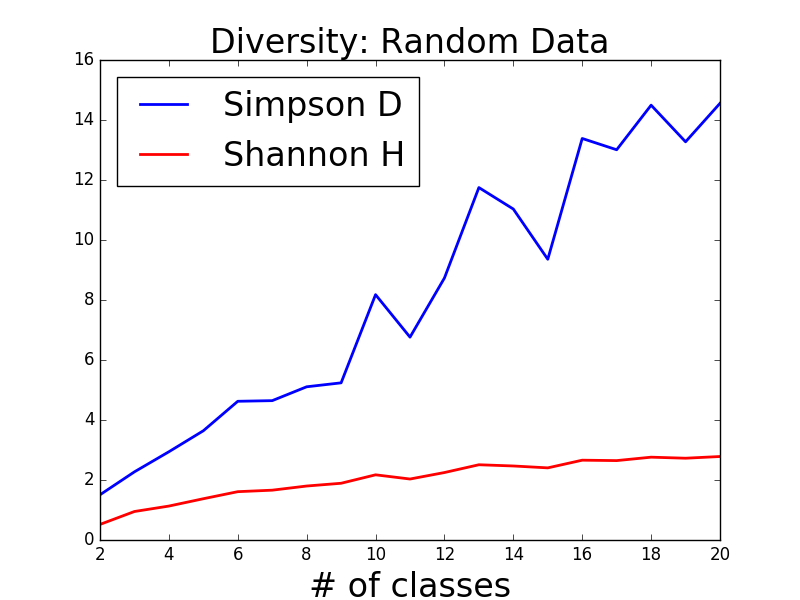} &
\includegraphics[width=0.23\linewidth]{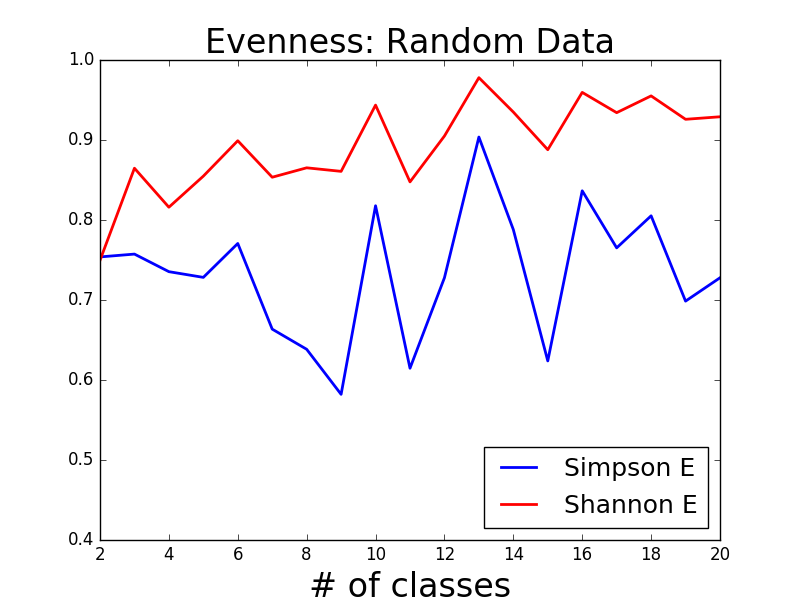} \\
(a) & (b) & (c) & (d)
\end{tabular}
\end{center}
\vspace{-0.2in}
\caption{Illustration of how (a) diversity and (b) evenness varies for a uniform distribution compared to how (c) diversity and (d) evenness varies for a random distribution.}
\label{fig:uni-rand}
\end{figure}

In this Section, we report on the statistical analysis of the ten facial coding schemes in the $DiF$ data set.   Intuitively, in order to provide sufficient coverage and balance, a data set needs to include data with diverse population characteristics.  This type of analysis comes up in multiple disciplines, including bio-diversity \cite{heip_1998,hill_ecology_73}, where an important objective is to quantify species diversity of ecological communities.  It has been reported that species diversity has two separate components: (1) species richness, or the number of species present, and (2) their relative abundances, called evenness. We use these same measures to quantify the diversity of face images using the ten facial coding schemes.  We compute diversity using Shannon $H$ and $E$ scores and Simpson $D$ and $E$ scores~\cite{heip_1998}. Additionally, we measure mean and variance for each of the feature dimensions of the ten facial coding schemes  The computation of diversity is as follows: given individual $p_i$ in a probability distribution for each feature measure, and the $S$ being the number of classes for the attribute, we compute:

\begin{figure}[h]
\begin{center}
\begin{tabular}{ll}
\hline
\bf Diversity   & \bf Evenness \\\hline\\
Shannon $H = -\sum_{i=1}^{S} p_i *ln (p_i)$         &Shannon $E = \frac{H}{ln(S)}$\\\\\hline\\
Simpson $D = \frac{1}{\sum_{i=1}^{S} (p_i*p_i)}$    &Simpson $E = \frac{D}{S}$\\\\\hline
\end{tabular}
\end{center}
\end{figure}
Shannon $H$ and Simpson $D$ are diversity measures and Shannon $E$ and Simpson $E$ are evenness measures. To see how they work, consider a $20$ class problem ($S=20$) with uniform distribution ($p_i=0.05$).  These measures take the following values: Shannon $H = 2.999$, Shannon $E = 1.0$, Simpson $D = 2.563$, and Simpson $E = 1.0$.  Evenness is constant at $1.0$ as expected.  Shannon $D$ represents the diversity of $20$ classes ($e^{2.999} \approx 20$).  For complex distributions, it may not be easy to understand the meaning of specific values of these scores.  Generally, a higher diversity value is better than a lower value, whereas an evenness value closer to $1.0$ is better.  Figure~\ref{fig:uni-rand} illustrates these measures on two example distributions. Figure~\ref{fig:uni-rand} (a) and (b) show how diversity and evenness values vary for a uniform distribution, respectively, as the number of classes increase from $2$ to $20$.   Figure~\ref{fig:uni-rand} (c) and (d) show the same information for a random distribution.

Table~\ref{tab:div} summarizes the diversity scores computed for the ten facial coding schemes in the $DiF$ data set. As described in Section~\ref{sec:codingschemes}, many of the coding schemes have multiple dimensions.  Hence the table has more than ten rows. The craniofacial measurements across the three coding scheme types total $28$ features corresponding to craniofacial distances, craniofacial areas and craniofacial ratios. The diversity scores of the different dimensions of the remaining seven coding schemes can similarly be seen in Table~\ref{tab:div}.  

\begin{table}[htp]
\begin{small}
\sisetup{round-precision=2}
\begin{tabular}{|m{0.9in}|m{1.2in}|P{0.6in}|P{0.6in}|P{0.60in}|P{0.60in}|c|c|}
\hline
{\bf Coding Scheme}&{\bf Measurement} &  {\bf Simp. $D$} & {\bf Simp. $E$} & {\bf Shan. $H$} & {\bf Shan. $E$} & {\bf Mean} & {\bf Var}\\ \hline \hline
\multirow{8}{0pt}{Craniofacial Distance} 
 & $n-sto$ & 5.875 & 0.979 & 1.781 & 0.994 & 33.20	& 3.93 \\ 
 &  $ps-pi$  & 5.886 & 0.981 & 1.782 & 0.995 & 3.26	& 0.98 \\ 
 &  $or-sci$ & 5.867 & 0.978 & 1.780 & 0.994  & 12.6 & 2.22  \\ 
 &  $sn-cc$ & 5.832 & 0.972 & 1.777 & 0.992  & 6.50	&  1.69\\ 
 & $sn-sto$ &  5.870 & 0.978 & 1.780 & 0.994  & 9.95	&  2.12\\ 
 & $sto-li$ &  5.862 & 0.977 &  1.780 & 0.994 & 5.86	& 1.97 \\ 
 &  $cph-cph$ & 5.872 & 0.979 & 1.781 & 0.994  & 6.99	& 1.35 \\ 
 &  $sbal-lss$ &  5.832 & 0.972 & 1.777 & 0.992  & 7.07	& 1.63 \\  \hline
 \multirow{12}{0pt}{Craniofacial Area}
 & $tr-n$ &  5.887 & 0.981 & 1.782 & 0.995 & 33.83	& 5.25 \\ 
 & $tr-gn$ &  5.879 & 0.980 & 1.781 & 0.994 & 89.27	& 9.44 \\ 
 & $n-gn$ &  5.877 & 0.979 & 1.781 & 0.994  & 55.44	&  7.48\\ 
 & $sn-gn$ &  5.866 & 0.978 & 1.780 & 0.993 & 32.19	&  5.82\\ 
 & $zy-zy$ & 5.868 & 0.978 & 1.780 & 0.994  & 63.67	& 4.42 \\ 
 & $go-go$ &  5.888 & 0.981 & 1.782 & 0.995 & 43.47	& 3.89 \\ 
 & $en-en$ & 5.875 &0.979 & 1.781 & 0.994  & 17.49	& 1.06 \\ 
 & $en-ex$ &  5.880 & 0.980 & 1.781 & 0.994 & 11.56	&  0.56\\ 
 & $ex-ex$ & 5.882 & 0.980 & 1.782 & 0.994  & 40.62	&  1.00\\ 
 & $n-sn$ &  5.873 & 0.979 & 1.781 & 0.994  & 23.24	&  3.02\\ 
 & $al-al$ & 5.876 & 0.979 &1.781 & 0.994 & 13.62	& 1.65 \\ 
& $ch-ch$ & 5.858 & 0.976 & 1.780 & 0.993 & 27.09	&  4.04\\  \hline
\multirow{8}{0pt}{Craniofacial Ratio} 
 & $(n-gn)/(zy-zy)$  &  5.878 & 0.980 & 1.781 & 0.994  & 0.87	& 0.11 \\
 & $(sto-gn)/(go-go)$  & 5.878 & 0.980 & 1.781 & 0.994  & 0.51	& 0.08\\ 
 & $(en-en)/(ex-ex)$ &  5.885 & 0.981 & 1.782 & 0.995 & 0.43	& 0.02 \\ 
 & $(ex-en)/(en-en)$ &  5.870 & 0.978 & 1.781 & 0.994 & 0.66	& 0.06 \\ 
 & $(ps-pi)/(ex-en)$ &  5.878 & 0.980 & 1.781 & 0.994 & 0.28	& 0.08 \\ 
 & $(al-al)/(n-sn)$ &  5.902 & 0.984 &1.783 & 0.995 & 0.59	&  0.08\\ 
 & $(ls-sto)/(sto-li)$  & 5.884 & 0.981 & 1.782 & 0.994  & 0.67	& 0.34 \\ 
 & $(ch-ch)/ (zy-zy)$ & 5.872 & 0.979 & 1.781 & 0.994 & 0.42& 0.06 \\
 \hline
\multirow{2}{0pt}{Facial Symmetry} 
 & Density difference & 5.510 & 0.918 & 1.748 & 0.975 & 0.12 & 0.07 \\
 & Edge or. similarity & 5.002 & 0.834 & 1.692 & 0.944 & 0.01 & 0.01\\ \hline
\multirow{9}{0pt}{Facial Contrast} 
& Lips $L$ contrast &  5.857 & 0.976 & 1.779 & 0.993 & -0.07 & 0.09\\
& Lips $a$ contrast &  5.780 & 0.963 & 1.772 & 0.989 & 0.02 & 0.02\\ 
& Lips $b$ contrast & 5.868 & 0.978 & 1.780 & 0.994  & -0.01 & 0.01\\ 
& Eyes $L$ contrast &  5.724 & 0.954 & 1.766 & 0.986 & -0.18 & 0.14\\ 
& Eyes $a$ contrast & 5.872 & 0.979 & 1.781 & 0.994 & -0.02 & 0.01\\ 
& Eyes $b$ contrast & 5.862 & 0.977 & 1.780 & 0.993  & -0.03 & 0.02\\ 
& Eb $L$ contrast &  5.722 & 0.954 & 1.766 & 0.986 & -0.11 & 0.11\\ 
& Eb $a$ contrast & 5.759 & 0.960 & 1.769 & 0.987 & -0.01 & 0.01\\ 
& Eb $b$ contrast & 5.666 & 0.944 & 1.760 & 0.982 & -0.01 & 0.01\\ \hline
Skin Color & ITA & 5.283 & 0.755 & 1.773 & 0.911  & 13.99 &  45.13 \\ \hline
Age & Age prediction &  4.368 & 0.624 & 1.601 & 0.823 & 26.28  & 14.65 \\ \hline
Gender & Gender prediction &  3.441 & 0.573 & 1.488 & 0.831  & 0.27 & 0.32\\ \hline
\multirow{2}{0pt}{Subjective Annotation}
  & Gender labeling & 2.000 & 1.000 & 0.693 & 1.000 & 0.49 & 0.50 \\
 & Age labeling &  4.400 & 0.629 & 1.675 & 0.861 &  30.44 & 16.99 \\
 \hline
 \multirow{3}{0pt}{Pose \& Resolution}
& Pose & 1.224 & 0.408 & 0.390 & 0.355 & 0.11 & 0.84\\ 
& IOD & 4.989 & 0.832 & 1.691 & 0.944 & 43.60 & 22.14\\ 
 & Face Region Size & 2.816 & 0.469 & 1.197 & 0.668 & 93.29 & 42.95\\ \hline
\end{tabular}
\end{small}
\caption{Summary of facial coding scheme analysis for the $DiF$ data set using Simpson $D$ (diversity), Simpson $E$ (evenness), Shannon $H$ (diversity), Shannon $E$ (evenness), mean and variance.}
\label{tab:div}
\end{table}

\subsection{Coding Scheme 1: Craniofacial Distances}

\begin{figure}[t]
\begin{center}
\includegraphics[width=\linewidth]{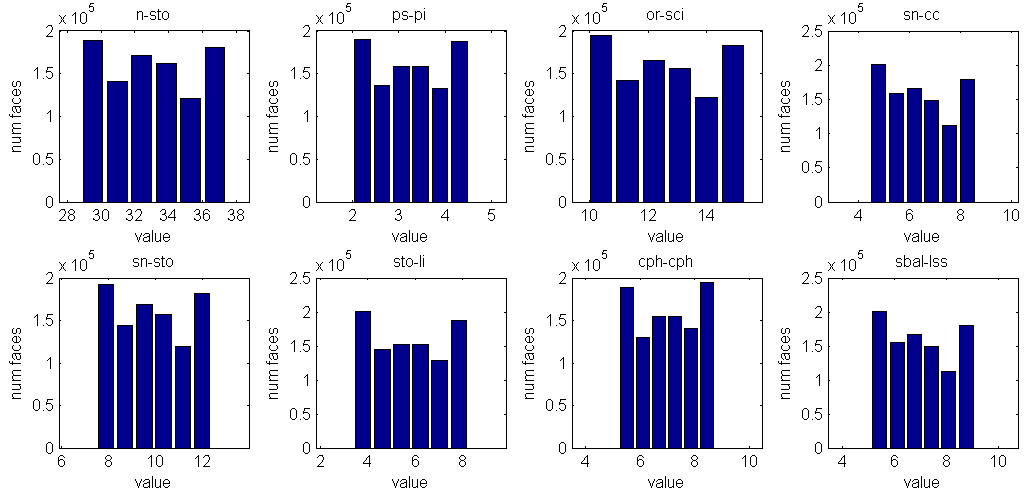} 
\end{center}
\vspace{-0.3cm}
\caption{Feature distribution of craniofacial distances (coding scheme 1) for the $DiF$ data set.}
\label{fig:cranio_distances}
\end{figure}

Figure \ref{fig:cranio_distances} summarizes the feature distribution for the $8$ craniofacial distances in coding scheme 1.  The highest Simpson $D$ value is $5.888$ and the lowest is $5.832$. The highest and lowest Shannon $H$ values are $1.782$ and $1.777$. Based on the Shannon $H$ values, this feature dimension would typically map to $6$ classes.  Evenness is generally balanced with highest Simpson $E$ and Shannon $E$ of $0.981$ and $0.995$, respectively.

\subsection{Coding Scheme 2: Craniofacial Areas}

\begin{figure}[t]
\begin{center}
\includegraphics[width=\linewidth]{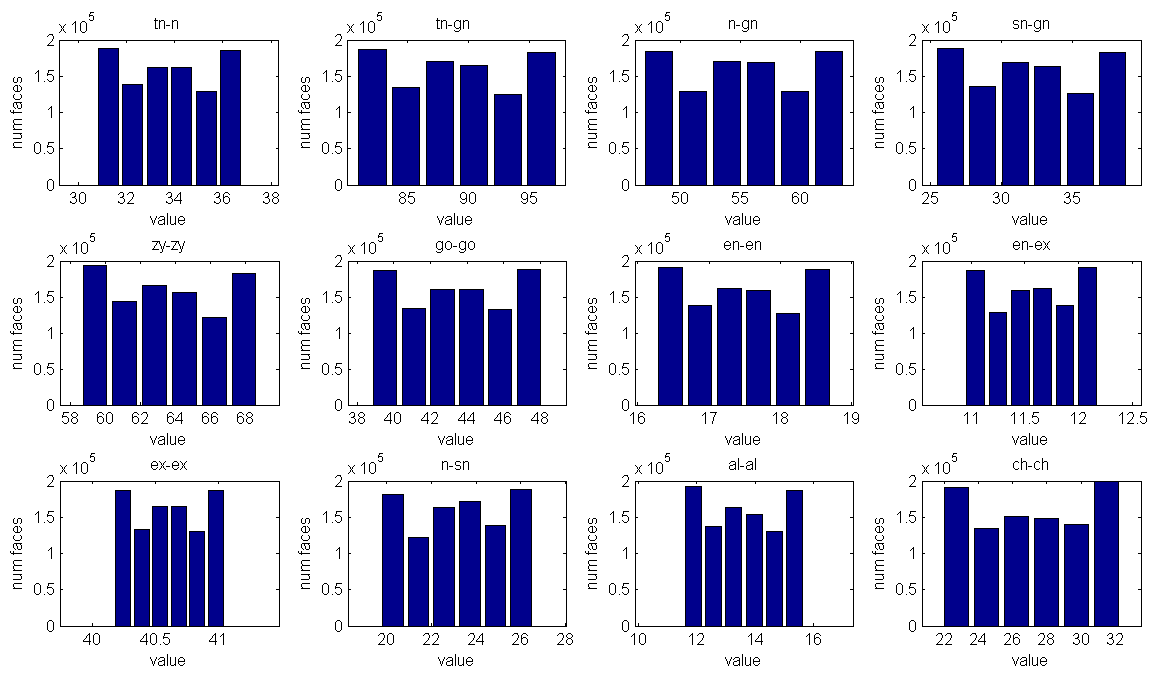} 
\end{center}
\vspace{-0.6cm}
\caption{Feature distribution of craniofacial areas (coding scheme 2) for the $DiF$ data set.}
\label{fig:cranio_areas}
\end{figure}

Figure \ref{fig:cranio_areas} summarizes the feature distribution for the $12$ craniofacial areas in coding scheme 2.  The highest Simpson $D$ value is $5.888$ and the smallest is $5.858$. The highest Shannon $H$ value is $1.782$ and the lowest is $1.780$. Compared to coding scheme 1, these values are in the similar range, mapping to $6$ classes. Evenness ranges between $0.981$ and $0.976$.  

\subsection{Coding Scheme 3: Craniofacial Ratios}

\begin{figure}[t]
\begin{center}
\includegraphics[width=\linewidth]{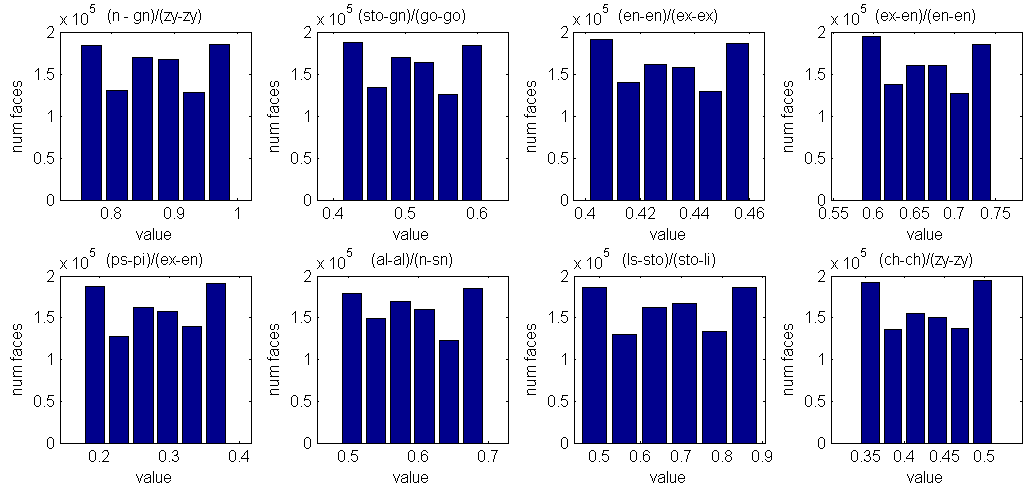} 
\end{center}
\vspace{-0.6cm}
\caption{Feature distribution of craniofacial ratios (coding scheme 3) for the $DiF$ data set.}
\label{fig:cranio_ratios}
\end{figure}

Figure \ref{fig:cranio_ratios} summarizes the feature distribution for the $8$ craniofacial ratios in coding scheme 3.  
The largest Simpson $D$ value is $5.902$ and smallest is $5.870$. Similarly, the largest Shannon $H$ value is $1.783$ and smallest is $1.781$.  This would map to approximately to $6$ classes. While Simpson $E$ has a range between $0.978$ to $0.984$, Shannon $E$ ranges between $0.994$ to $0.995$. The evenness of coding scheme 3 is similar to coding scheme 2. 

\subsection{Coding Scheme 4: Facial Symmetry}

\begin{figure}[htbp]
\begin{center}
\begin{tabular}{cc}
\includegraphics[width=0.3\linewidth]{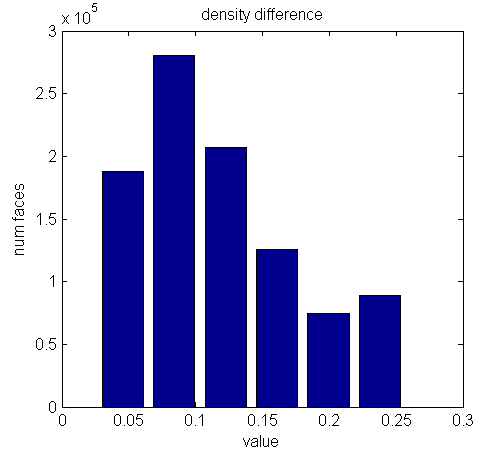} &
\includegraphics[width=0.31\linewidth]{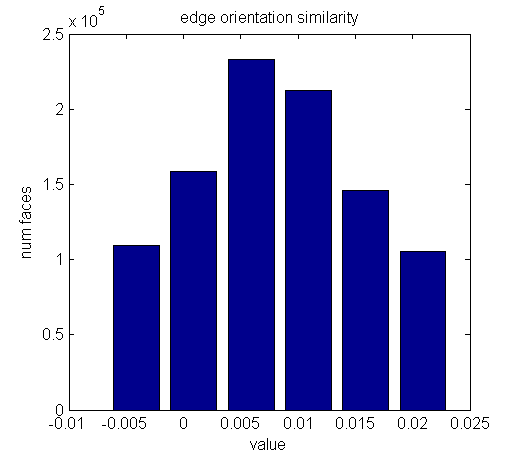} \\
(a) & (b)
\end{tabular}
\end{center}
\vspace{-0.6cm}
\caption{Feature distribution of facial symmetry (coding scheme 4):  (a) density difference and (b) edge orientation similarity for the $DiF$ data set.}
\label{fig:symmetry_distribution}
\end{figure}

Figure \ref{fig:symmetry_distribution} summarizes the feature distribution for facial symmetry in coding scheme 4.  The diversity value is in a middle range compared to the previous coding schemes. For example, the highest Simpson $D$ is $5.510$ and the largest Shannon $H$ is $1.748$. The evenness values are lower as well with highest Simpson $E$ value being $0.918$ and highest Shannon $E$ value being $0.975$. The Shannon $H$ value of $1.692$ translates to about $5.4$ classes. 

\subsection{Coding Scheme 5: Facial Regions Contrast}
\begin{figure}[htbp]
\begin{center}
\includegraphics[width=\linewidth]{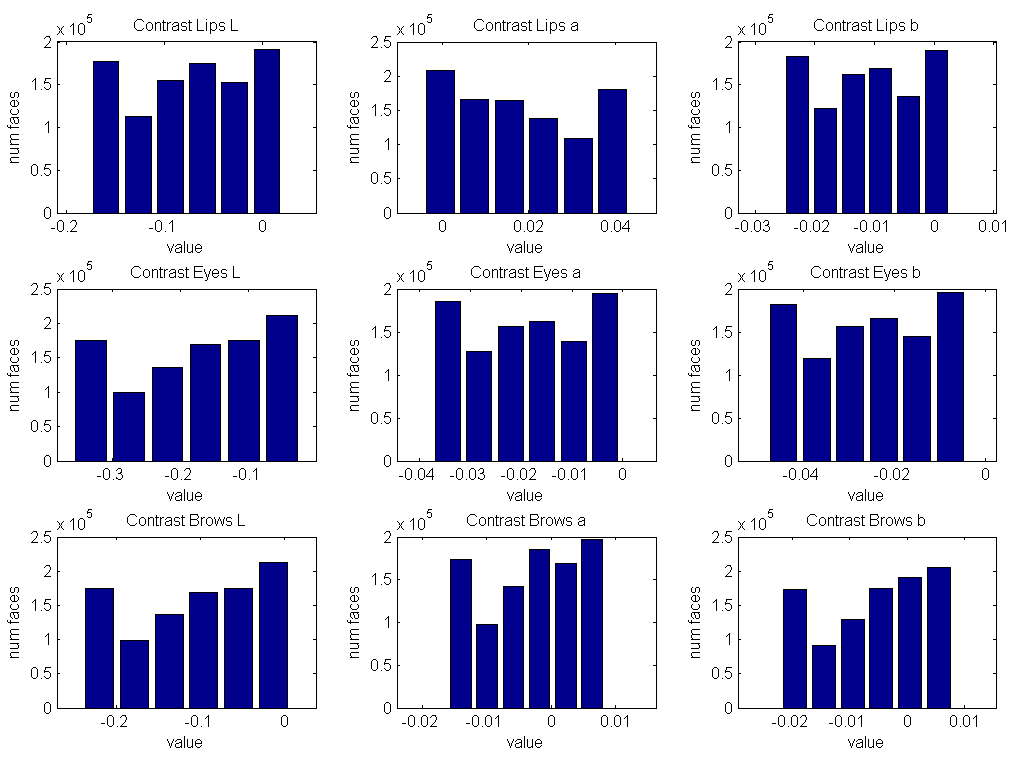} 
\end{center}
\vspace{-0.3cm}
\caption{Feature distribution of facial regions contrast (coding scheme 5) for the $DiF$ data set.}
\label{fig:contrast_dist}
\end{figure}

Figure \ref{fig:contrast_dist} summarizes the feature distribution for facial contrast in coding scheme 5.  The highest Simpson $D$ value is $5.872$ and highest Shannon $H$ value is $1.781$, which is equivalent to $5.9$ classes. The evenness factor Shannon $E$ is very close to $0.979$ indicating that the measures are close to even. 

\subsection{Coding Scheme 6: Skin Color}

\begin{figure}[htbp]
\begin{center}
\includegraphics[width=0.3\linewidth]{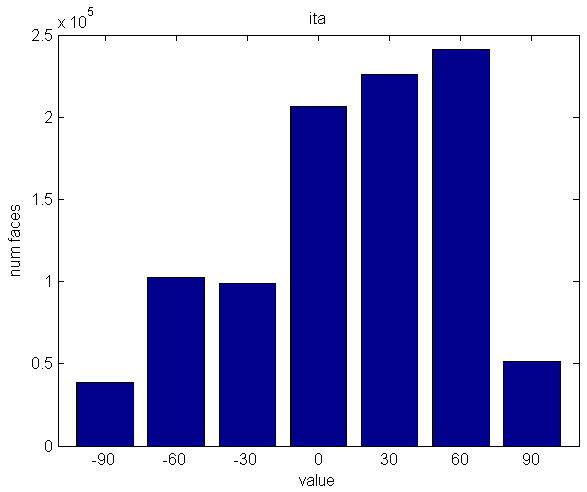} 
\end{center}
\vspace{-0.6cm}
\caption{Feature distribution of skin color using Individual Typology Angle (ITA) (coding scheme 6) for the $DiF$ data set.}
\label{fig:ita_dist}
\end{figure}

Figure \ref{fig:ita_dist} summarizes the feature distribution for skin color in coding scheme 6.  The Simpson $D$ value is $5.283$ and Shannon $H$ value is $1.773$ which translates to about $5.88$ classes, which shows a good match with the number of bins we used. The evenness is weaker than a uniform distribution. 

\subsection{Coding Scheme 7: Age Prediction}

\begin{figure}[htbp]
\begin{center}
\begin{tabular}{cc}
\includegraphics[width=0.3\linewidth]{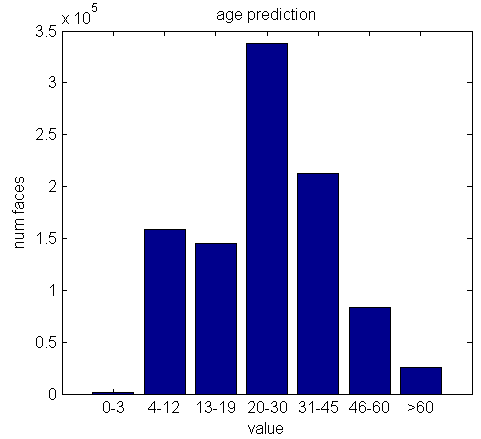} &
\includegraphics[width=0.3\linewidth]{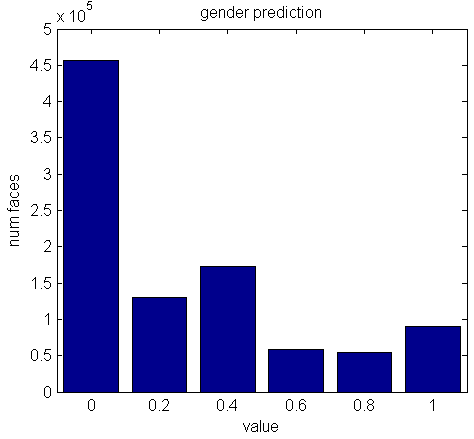} \\
(a) & (b)
\end{tabular}
\end{center}
\vspace{-0.6cm}
\caption{Feature distribution of (a) age prediction (coding scheme 7) and (b) gender prediction (coding scheme 8) for the $DiF$ data set.}
\label{fig:gen_div}
\end{figure}

Figure \ref{fig:gen_div}(a) summarizes the feature distribution for age prediction in coding scheme 7, where we bin the age values into seven groups: [0-3],[4-12],[13-19],[20-30],[31-45],[46-60],[61-].  The Simpson $D$ and Shannon $H$ values are $4.368$ and $1.601$. Because of the data distribution not being even, we can see a lower $E$ value around $0.624$. The Shannon $H$ value of $1.601$ maps to $5$ classes.

\subsection{Coding Scheme 8: Gender Prediction}

Figure \ref{fig:gen_div} also summarizes the feature distribution for gender prediction in coding scheme 8.  Even though this has two classes, male and female, the confidence score ranges between $0-1$.  The gender score distribution is shown in  Figure~\ref{fig:gen_div} (b). The Simpson $D$ is $3.441$ and Shannon $H$ is $1.488$. The Shannon $H$ value translates to $4.4$ classes, which is beyond the typical two classes used for gender, possibly reflecting the presence of sub-classes. The Simpson evenness score of $0.573$ reflect some unevenness as well. 

\subsection{Coding Scheme 9: Subjective Annotation}

\begin{figure}[t]
\begin{center}
\begin{tabular}{cc}
\includegraphics[width=0.3\linewidth]{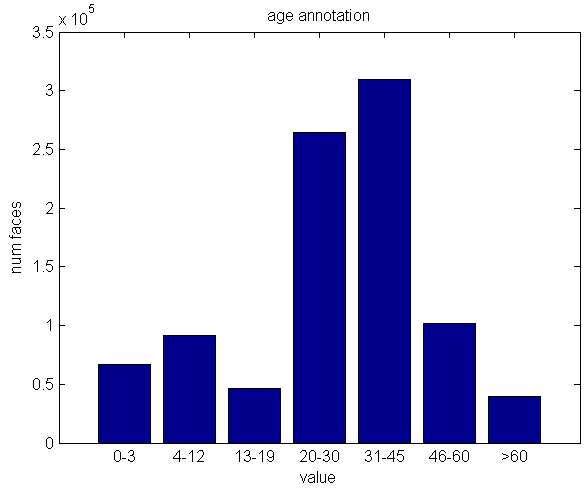} &
\includegraphics[width=0.3\linewidth]{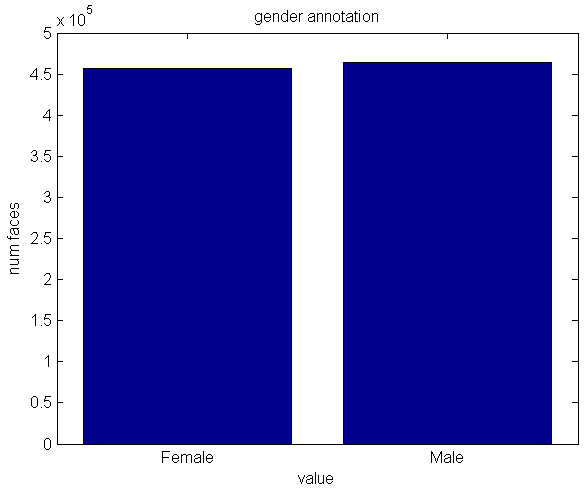}
 \\
(a) & (b)
\end{tabular}
\end{center}
\vspace{-0.6cm}
\caption{Feature distributions of subjective annotations (coding scheme 9) for (a) age and (b) gender for the $DiF$ data set.}
\label{fig:human_annotation_distribution}
\end{figure}

Figure \ref{fig:human_annotation_distribution} summarizes the feature distribution for the subjective annotations of age and gender for coding scheme 9.  The Simpson $D$ for gender distribution is $2.0$ and Shannon $H$ is $0.693$, indicating the equivalent classes to be near $2$, which is understandable. The evenness is very high, indicating a nearly flat distribution.  The Simpson $D$ is $4.368$ and Shannon $H$ is $1.675$, resulting in a equivalent class index of approximately $5.3$. However, the evenness scores are low at $0.629$, indicating unevenness, as is visible in the distribution of the annotated age scores.  

\subsection{Coding Scheme 10: Pose and Resolution}

\begin{figure*}[htbp]
\begin{center}
\begin{tabular}{ccc}
\includegraphics[width=0.3\linewidth]{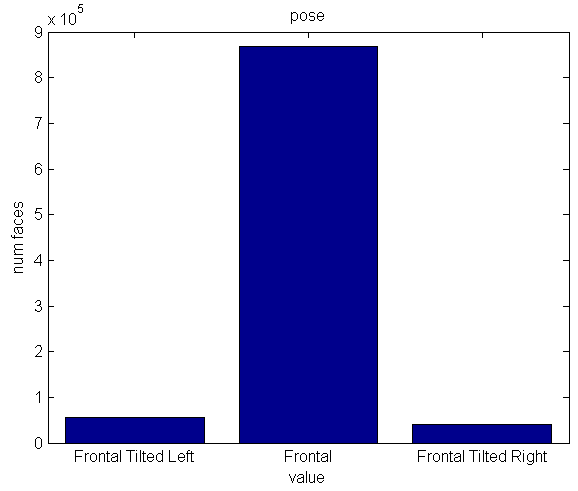} &  \includegraphics[width=0.3\linewidth]{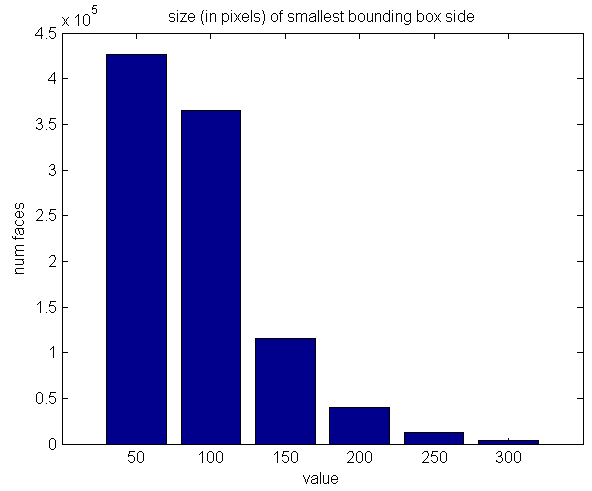} &
\includegraphics[width=0.3\linewidth]{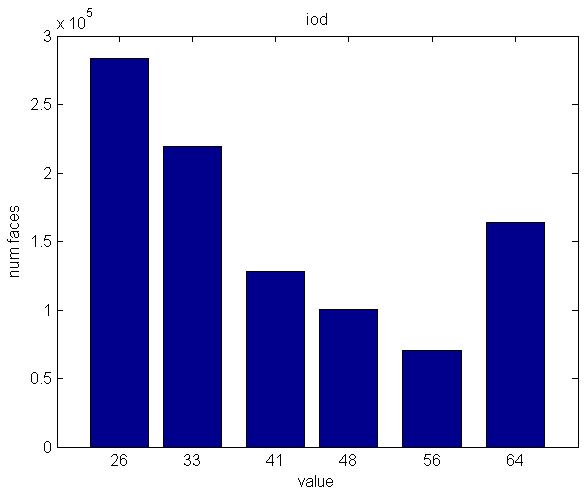} \\
(a) & (b) & (c)
\end{tabular}
\end{center}
\vspace{-0.6cm}
\caption{Feature distribution of pose and resolution (coding scheme 10) for the $DiF$ data set, including (a) pose, (b) face region bounding box size, (c) intra-ocular distance (IOD).}
\label{fig:resolution}
\end{figure*}

Figure \ref{fig:resolution} summarizes the feature distribution for pose and resolution for coding scheme 10.  Pose uses three dimensions from the output of DLIB face detection and the distribution is shown in ~\ref{fig:resolution} (a). When computing mean and variance for pose in Table \ref{tab:div}, we used the following values: Frontal Tilted Left -1, Frontal 0, and Frontal Tilted Right 1. The IOD and box size distribution are shown in Figure~\ref{fig:resolution} (b)-(c). The distances have been binned to six classes. The three class pose distribution has a Shannon $H$ value of $0.39$. The Shannon $H$ value for IOD is $1.69$ (mapping to equivalent of $5.4$ classes) while for the box size it is $1.197$, translating to $3.3$ classes.  

\subsection{Discussion}

Some observations come from this statistical analysis of the ten coding schemes on the $DiF$ face image data.  One is that the many of the dimensions of the craniofacial schemes have high scores in diversity relative to the other coding schemes.  Generally, they are higher than measures used for age and gender, whether using a predictive model or subjective annotation.  Similarly, their evenness scores are also closer to one.  What this shows is that there is higher variability in these measures, and they are capturing information that age and gender alone do not.  Interestingly, facial regions contrast, which was designed to capture information about age, has a higher diversity score and better evenness than that for either neural network prediction of age or subjective human annotation of age.  Again, it implies that this continuous valued feature of facial contrast is capturing information that goes beyond simple age prediction or labeling.   The only feature dimension with lower diversity is understandably pose, which was a controlled variable in selecting images for the $DiF$ data set, since only mostly frontal faces were incorporated.  In future work, we will use these methods to assess diversity of other face data sets, which will provide a basis for comparison and further insight.



\section{Summary and Future Work} \label{sec:conclusion}

We described the new Diversity in Faces ($DiF$) data set, which has been developed to help advance the study of fairness and accuracy in face recognition technology.  $DiF$ provides a data set of annotations of publicly available face images sampled from the YFCC-100M data set of 100 million images.  The annotations are defined from facial coding schemes that provide quantitative measures related to intrinsic characteristics of faces including craniofacial features, facial symmetry, facial contrast, skin color, age, gender, subjective annotations and pose and resolution.  We described the process for generating the $DiF$ data set as well as the implementation and extraction of the ten facial coding schemes.  We also provided a statistical analysis of the facial coding scheme measures on the one million $DiF$ images using measures of diversity,  evenness and variance.  For one, this kind of analysis has provided insight into how the $47$ total feature dimensions within the ten facial coding schemes provide measures of data set diversity for the one million images.  While it may not yet be possible to conclude that the goal is to drive all of these feature dimensions to be maximally diverse and even, we believe the approach outlined in this work provides a needed methodology for advancing the study of diversity for face recognition.   

There are multiple next directions for this work.  Table~\ref{tab:face_db} outlined many of the currently used face data sets.  We plan to perform the equivalent statistical analysis on some of these other data sets using the ten coding schemes.  This will provide an important basis for comparing data sets in terms of diversity.  Using the statistical measures outlined in this paper, including diversity, evenness and variance, we will begin to answer questions of whether one data set is better than another, or where a data set falls short in terms of coverage and balance.  We also strongly encourage others to build on this work.  We selected a solid starting point by using over one million publicly available face images and by implementing ten facial coding schemes.  We hope that others will find ways to grow the data set to include more faces.  As more insight comes from the type of analysis outlined in this paper, we see that an iterative process can more proactive sampling to fill in gaps.  For example, as technologies like Generative Adversarial Networks (GANs) continue to improve~\cite{progressiveGAN_ICLR18,faceid_gan18}, it may be possible to generate faces of any variety to synthesize training data as needed.  We expect that $DiF$ will be valuable to steer these generative methods towards gaps and blind spots in the training data, for example, to assist methods that automatically generate faces corresponding to a specific geographical area~\cite{bessinger_wacv19}.   We also hope that others will see ways to improve on the initial ten coding schemes and add new ones.  Pulling together our collective efforts is the best way to make progress on this important topic.  We hope that the $DiF$ data set provides a useful foundation for creating more fair and accurate face recognition systems in practice.
\clearpage
\bibliographystyle{unsrt}
\bibliography{dif}
\end{document}